\documentclass[lettersize,journal]{IEEEtran}
\usepackage{amsmath,amsfonts}
\usepackage{algorithmic}
\usepackage{algorithm}
\usepackage{array}
\usepackage[caption=false,font=normalsize,labelfont=sf,textfont=sf]{subfig}
\usepackage{textcomp}
\usepackage{stfloats}
\usepackage{url}
\usepackage{verbatim}
\usepackage{graphicx}
\usepackage{cite}
\usepackage{booktabs}

\usepackage{doi}
\begin{document}
\bibliographystyle{IEEEtran}
\title{SA-DNet: A on-demand semantic object registration network adapting to non-rigid deformation}

\author{Housheng Xie, Junhui Qiu, Yuan Dai, Yang Yang,~\IEEEmembership{Member,~IEEE,} Changcheng Xiang and Yukuan Zhang
	\thanks{Housheng Xie, Junhui Qiu, Yang Yang and Yukuan Zhang are with the Laboratory of Pattern Recognition and Artificial Intelligence, Yunnan Normal University, Kunming, 650500, China (e-mail: xiehoushengsc@163.com; QiuJHaca@163.com; yyang$\textunderscore$ynu@163.com; Ykuan$\textunderscore$Zhang@163.com).}
	\thanks{Yuan Dai is with the Faculty of Science $\&$ Engineering, University of Nottingham Ningbo China, Ningbo 315000, China (e-mail: vesdy0328@gmail.com).}
	\thanks{Changcheng Xiang is with the Aba Teachers University, Aba 623002, China (e-mail: 810225636@qq.com).}
	\thanks{Corresponding authors: Yang Yang; Changcheng Xiang.}}

\markboth{}%
{Shell \MakeLowercase{\textit{et al.}}: A Sample Article Using IEEEtran.cls for IEEE Journals}

\markboth{}%
{Shell \MakeLowercase{\textit{et al.}}: A Sample Article Using IEEEtran.cls for IEEE Journals}


\maketitle

\begin{abstract}
As an essential processing step before the fusing of infrared and visible images, the performance of image registration determines whether the two images can be fused at correct spatial position. In the actual scenario, the varied imaging devices may lead to a change in perspective or time gap between shots, making significant non-rigid spatial relationship in infrared and visible images. Even if a large number of feature points are matched, the registration accuracy may still be inadequate, affecting the result of image fusion and other vision tasks. To alleviate this problem, we propose a Semantic-Aware on-Demand registration network (SA-DNet), which mainly purpose is to confine the feature matching process to the semantic region of interest (sROI) by designing semantic-aware module (SAM) and HOL-Deep hybrid matching module (HDM). After utilizing TPS to transform infrared and visible images based on the corresponding feature points in sROI, the registered images are fused using image fusion module (IFM) to achieve a fully functional registration and fusion network. Moreover, we point out that for different demands, this type of approach allows us to select semantic objects for feature matching as needed and accomplishes task-specific registration based on specific requirements. To demonstrate the robustness of SA-DNet for non-rigid distortions, we conduct extensive experiments by comparing SA-DNet with five state-of-the-art infrared and visible image feature matching methods, and the experimental results show that our method adapts better to the presence of non-rigid distortions in the images and provides semantically well-registered images.
\end{abstract}


\section{Introduction}
\IEEEPARstart{I}{mage} registration, in its broad definition, seeks to find a good correspondence of the identical structure or semantics between two images. Many computer vision tasks rely on establishing accurate correspondence, such as image fusion\cite{li2021rfn}\cite{tang2022image}\cite{zhou2021semantic}, visual Simultaneous Localization and Mapping (SLAM)\cite{shin2021self}\cite{yang2020robust}, Structure from Motion (SfM)\cite{lindenberger2021pixel}, change detection\cite{khelifi2020deep}\cite{chen2021remote}, etc. For infrared and visible image fusion task, the images are required to be registered before to fusion, then the same spatial position of pixels in the images are fused to generate an individual fused image that retains salient objective and texture detail information from the infrared and visible images. In other words, the semantic information in the fused image is better represented and facilitates subsequent vision tasks.
\begin{figure}[htbp]
        \centering
        \captionsetup[subfloat]{labelsep=none,format=plain,labelformat=empty}
        \subfloat [\rmfamily (a) Image fusion after registration by MatchFormer\cite{wang2022matchformer}]{
        \begin{minipage}{1\linewidth}
                \centering
                \includegraphics[width=1\linewidth]{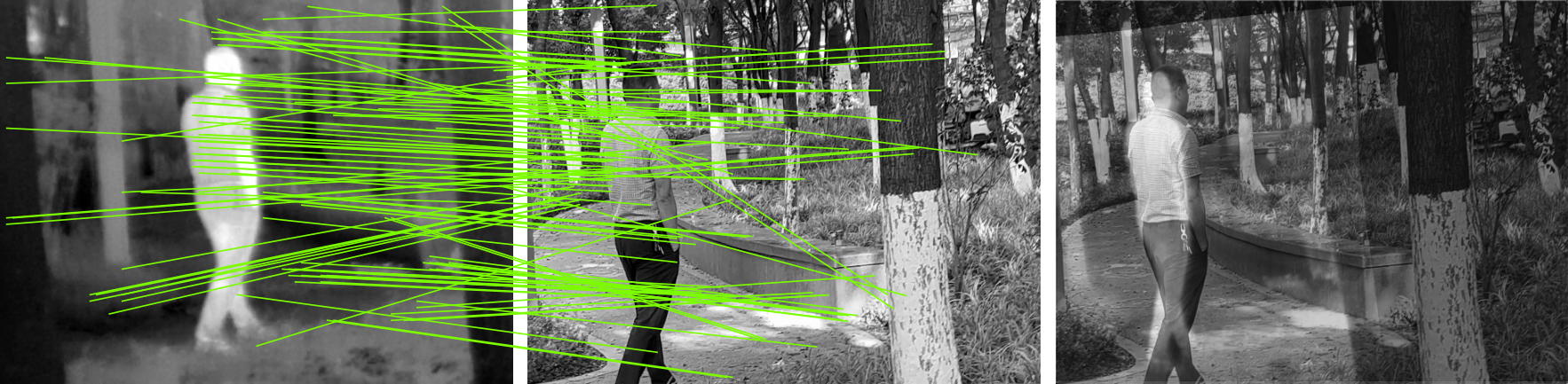}
                
                \label{chutian1}
        \end{minipage}
        }
        \qquad
        \subfloat [\rmfamily (b) Registration and fusion using SA-DNet]{
        \begin{minipage}{1\linewidth}
                \centering
                \includegraphics[width=1\linewidth]{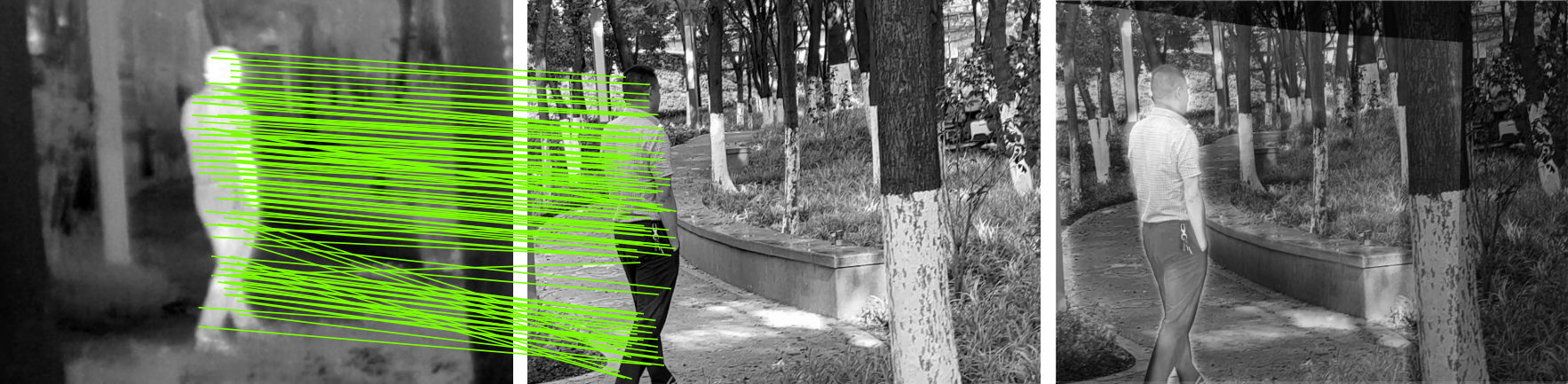}
               
                \label{chutian2}
        \end{minipage}
        }
               
        \caption{\textbf{Comparison between conventional registration and semantic-aware registration.} This example shows that SA-DNet is capable of withstanding the non-rigid distortion in the images and delivering better registered images for image fusion.}
\end{figure}

Due to the modal differences between infrared and visible images, perfectly registering the two images is not an easy work. After years of effort by researchers, many methods have been proposed to attempt to solve the challenge in registration of infrared and visible images, which broadly classified into two types: area-based and feature-based. The area-based methods first assume the transformation model between the images and then alternate two steps: similarity metric and transformation model parameter optimization. The similarity metric is used to quantify the image registration accuracy and guide the whole optimization process, which is usually designed based on the structural relationship between infrared and visible images. Area-based methods have the limitations of slow iteration speed, less robust to image noise and non-rigid transformations, so our work focuses on feature-based registration methods.

The feature-based methods typically consist three steps: feature extraction and description, feature matching, estimation of transformation model parameters. Due to the severe nonlinear intensity differences between infrared and visible images, traditional feature matching descriptor such as SIFT\cite{lowe2004distinctive}, ORB\cite{rublee2011orb}, and SURF\cite{bay2006surf} perform poorly under multimodal conditions. Therefore the researchers proposed UR-SIFT\cite{sedaghat2011uniform}, MSPC\cite{liu2018novel}, RIFT\cite{li2019rift} and other descriptors for infrared and visible feature matching. In addition,  deep learning based feature matching methods such as SuperPoint\cite{detone2018superpoint}+SuperGlue\cite{sarlin2020superglue}, LoFTR\cite{sun2021loftr}, MatchFormer\cite{wang2022matchformer}, have shown high generalizability for infrared and visible feature matching. 

Once the corresponding feature points have been obtained by the aforementioned methods, the transformation model is usually interpret the geometric relationship between the image pairs, warping and resampling the images to get the registered images. However, for the image fusion task shown in Fig. 1 (a), the infrared and visible images were collected by different imaging devices, bringing different variations in imaging perspective. Even with many correctly corresponding feature points, transformation models such as Homography and TPS are incapable of producing a satisfactory registration. Apparently, the significant non-rigid distortions between the infrared and visible images make it difficult to register every spatial position of the images using existing image registration methods, resulting in an unacceptable visual result for the final fused image. We explain the causes behind this observation based on extensive experimentation as follows.

\textbf{(i)} \emph{Inability to register severe distortion images adequately.} Existing methods of image registration use accurate corresponding feature points to establish the transformation model parameters that are image globally optimum for the intended uses, which include natural vision, remote sensing, and medical. However, such registration procedures generate substantial registration errors in local regions of the image when there are multiple non-rigid distortions present. Subsequent high-level vision tasks of image registration, such as fusion and detection, typically pay more attention to the local semantic objects in the image. Since there is an irreconcilable conflict between the registration error in the local area and the attention to local semantic regions in advanced vision tasks, it is necessary to improve the registration accuracy for the local area of the image under the current conditions in order to provide more suitable registration images for advanced vision tasks.

\textbf{(ii)} \emph{Insufficient number of reliable corresponding feature points can be achieved.} A sufficient number of corresponding feature points is necessary for obtaining properly registered images, nevertheless, owing to the different camera internal parameters, there is an inherent non-rigid deformation between the infrared and visible images. In addition, the variation in capture perspective and modal differences between infrared and visible images may cause unpredictable deformation in local areas of the images, making it difficult for existing methods to obtain sufficient reliable corresponding feature points in some complex scenes, especially when the matching area is restricted to a specific region. To improve the feature matching performance of infrared and visible images, additional feature information needs to be included.

\textbf{(iii)} \emph{Low robustness of single transformation model to local non-rigid deformation.} The accuracy of image registration still has a lot of room for improvement in comparison to the developed state of computer vision tasks like semantic segmentation and object detection. There has been little dedicated study on transformation models in image registration, and most image registration tasks use relatively fixed transformation models. When dealing with various non-rigid distortions, a single transformation model cannot provide proper registration. Optical flow-based methods have made great strides in using deep learning to estimate the deformation field, however, the nonlinear intensity differences between infrared and visible images make the application of such methods to be investigated. As a result, it is important to consider how to propose a registration method that can adapt to complex nonrigid distortion scenes using preexisting generic transformation models like TPS\cite{bookstein1991thin}.
\begin{figure}[htbp]
\centering
\includegraphics[width=90mm]{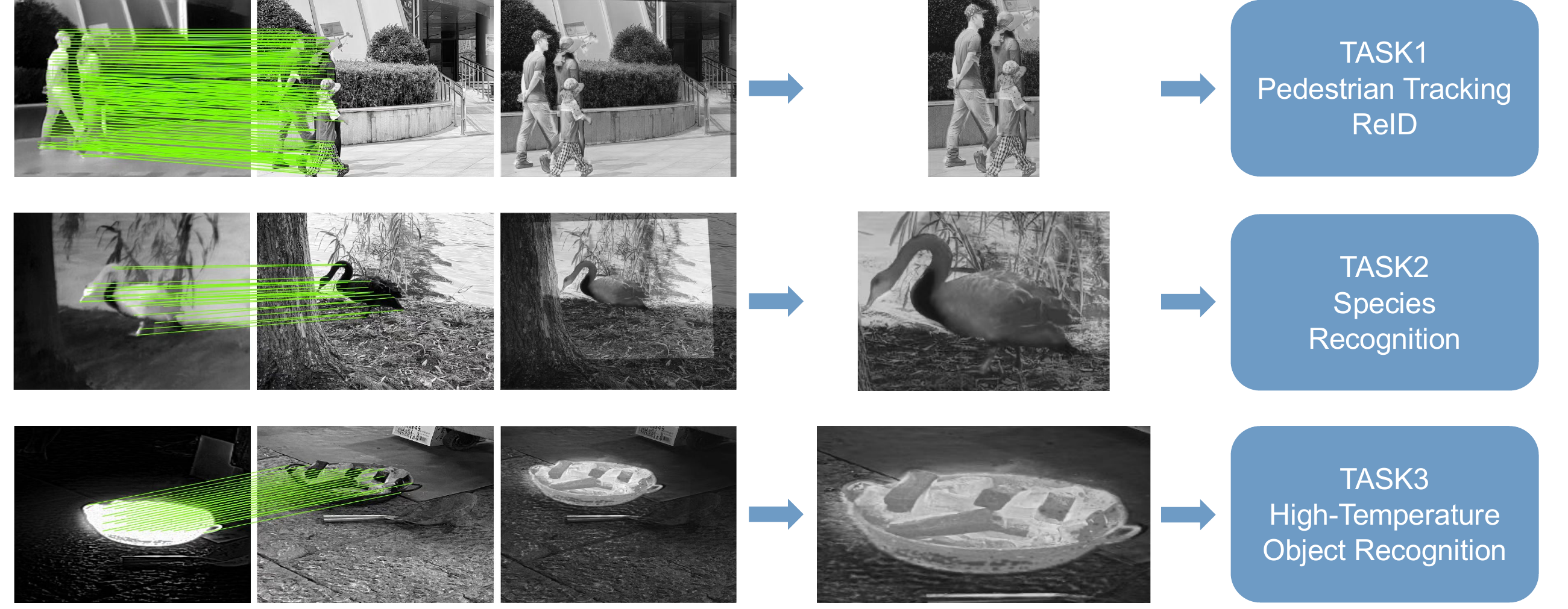}
\caption{\textbf{A description of semantic-aware on-demand registration.} For different visual tasks, various semantic objects can be selected for registration, and the semantic objects in the registration image can be precisely fused to facilitate the subsequent tasks.}
\label{fig2:env}
\end{figure}

To solve the above problems, we propose a Semantic-Aware on-Demand registration network (SA-DNet) for infrared and visible images. SA-DNet performs regionalized feature matching for semantic region of interest (sROI) in the images and computes TPS transformation with the corresponding feature points in sROI, which is illustrated in Fig. 1 (b). This approach enables the transformation to be optimally registered for the sROI and its adjacent area, the objects in sROI can correctly represent its complementary feature information after image fusion. On the other hand, as demonstrated in Fig. 2, the semantic-aware on-demand registration method yields well-registered sROI when adequate semantic-aware training data with mask annotation is available and allows for the selection of various objects as sROI for varied task needs. The key contributions of this paper can be summarized as follows:

1) We propose the concept of semantic aware on-demand registration, which enables semantic object and adjacent area to be accurately registered and facilitate subsequent vision tasks when there are numerous non-rigid distortions between infrared and visible images.

2) We design Hierarchical Orientation Line (HOL) Descriptors and Gaussian-weighted decay strategy to enhance feature matching inside sROI based on the edge information and deep feature of the semantic object.

3) We design a hybrid attention mechanism based on Local Window Self-Attention and Attention-Enhancing Strip Pool, which can efficiently extract and mix local and global features to achieve accurate perception of semantic objects.

\section{Related Work}
In this section, we review some references for solving infrared and visible images registration problems, as well as the transformation models applicable after feature matching.

\textbf{Feature-based matching of infrared and visible images} The differences between modals have led to the poor performance of traditional feature matching algorithms such as SIFT\cite{lowe2004distinctive}, ORB\cite{rublee2011orb}, and SURF\cite{bay2006surf} in infrared and visible images, so many researchers worked on developing a well-performing algorithm that can be used in infrared and visible image registration. Since PC\cite{morrone1987feature} is efficient in extracting information about the structure of images, the method of constructing feature descriptors by combining PC in the frequency domain is widely used. Ye et al\cite{ye2017robust} used a combination of PC and directional histogram to describe the feature points in terms of the structural properties in the images. Li et al\cite{li2019rift} proposed radiation invariant feature transform (RIFT) which combining PC and maximum index mapping (MIM), achieved good results in feature matching for various modalities. In RIFT, the authors first use FAST\cite{rosten2006machine} method identify duplicate corner and edge feature points in PC maps. As a further step, MIM obtains feature descriptor based on log-Gabor convolution sequence and by constructing multiple angles of MIM to achieve rotational invariance. Cui et al\cite{cui2019novel} constructed a scale space based on a nonlinear diffusion function and constructed a rotation descriptor to improve the effect of RIFT\cite{li2019rift}. However, these traditional methods rely only on local features for matching and are unable to interact with remote feature information to produce reliable correspondences in regions with less feature information.

The development of deep learning technology leads to better performance in infrared and visible image registration. Using Siamese CNN and non-weight sharing CNNs, \cite{baruch2021joint} merge joint and disjoint cues in infrared and visible images. \cite{ma2019novel} proposed a two-step registration method based on deep local feature information. The first step is to compute the approximate spatial relationships obtained by the pre-trained VGG network and use the deep features for matching. The second step applies the spatially-aware matching approach to the local feature, adjusting the results of the first step with more precise positional information. In addition, many methods such as D2-Net\cite{dusmanu2019d2}, R2D2\cite{revaud2019r2d2}, COTR\cite{jiang2021cotr}, and LoFTR\cite{sun2021loftr} have used SfM datasets to construct dense correspondence supervised training in recent years, bringing a new level of effectiveness for feature matching. D2-Net obtains valid keypoints by detecting local maxima on the feature map using a single CNN to simultaneously perform feature detection and extract dense feature descriptors. R2D2 thinks that repeatable points of interest may not have high discrimination and adds an additional branch for the prediction of the reliability map. SuperGlue\cite{sarlin2020superglue} describes the feature matching problem as quadratic assignment problems, and uses graph neural networks and attention mechanisms to achieve efficient feature matching. COTR, LoFTR and other methods, introducing Transformer\cite{vaswani2017attention} to interact with remote spatial feature information and constructing more discriminative descriptors. MatchFormer\cite{wang2022matchformer} replaces the ResNet backbone network in LoFTR using Self-Attention and Cross-Attention and achieve more efficient matching.

\textbf{Transformation} Transformation models mainly consist mostly of linear and nonlinear models, with rotation, etc. being the simplest linear models. In image registration, the Homography model is often used, which presupposes a flat or nearly flat scene. However, most images of natural scenes are not lie on the same plane, if we know the camera internal parameters, we can get a more accurate transformation model. The SLAM task usually combining internal camera parameters to interprets the transformation relationship of the images using the fundamental matrix, which constrains the points in the image to the lines in another image and reflects the projective geometry relationship of epipolar geometry.

The nonlinear model is primarily dealing with the scenario in which the image has several non-rigid distortions. In image registration, the non-rigid transformation model is often a geometric transformation based on interpolation or approximation theory. The thin-plate spline function (TPS)\cite{bookstein1991thin} is a frequently used interpolation method, and typically based on 2D interpolation, which is often used in image registration. The TPS deforms these N points to their appropriate locations while deforming the whole space. In free form deformation (FFD)\cite{sederberg1986free}, the image is divided into a number of rectangular units before being warped under the effect of control points using cubic B splines. Then the FFD uses the position of the grid points to calculate the coordinate offset of each pixel point, which finally resamples the image pixel by the coordinate offset to achieve its non-rigid transformation.
\begin{figure*}[htbp]
\centering
\includegraphics[width=160mm]{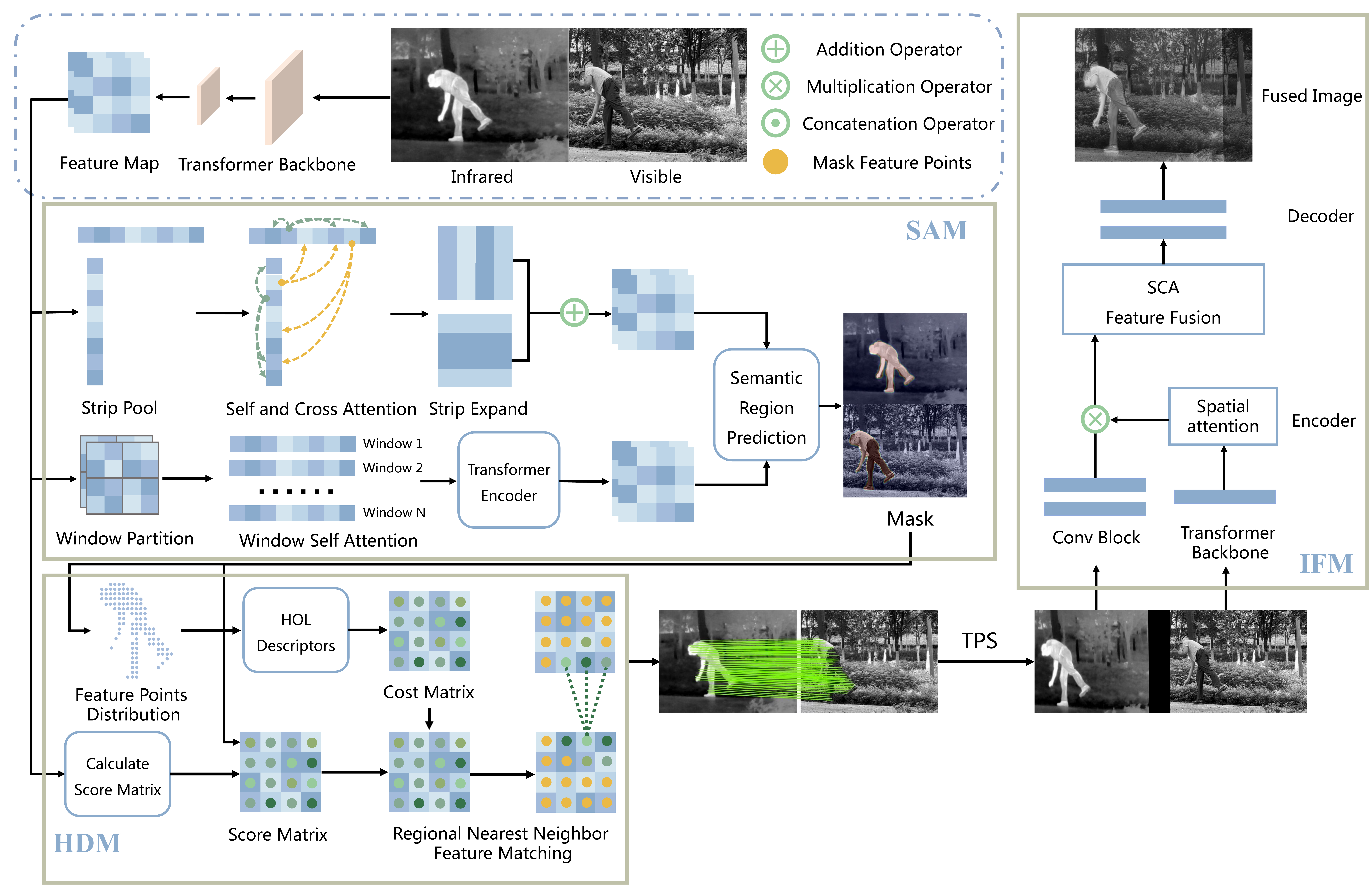}
\caption{\textbf{Overview of SA-DNet structure.} Specifically, Transformer backbone extracts feature maps for infrared and visible images, Local Window Self-Attention and Attention-Enhancing Strip Pool extract the sROI in the images based on the feature maps. HDM performs regional nearest neighbor feature matching inside sROI by combining HOL and deep feature information, the TPS model transforms the images based on the corresponding feature points, and the registered images will be used for image fusion.}
\label{fig2:env}
\end{figure*}

\section{Methodology}
Unlike previous feature matching methods\cite{fan2022seeing}\cite{sun2021loftr}\cite{wang2022matchformer}\cite{revaud2019r2d2}\cite{ma2019novel}\cite{baruch2021joint}, SA-DNet performs feature matching in the semantic region of interest (sROI). As an attempt, we hope that this kind of image registration method, known as semantic-aware on-demand registration, can be a new trend in registration method and provide a certain degree of solution to the registration problem in the scenarios with significant non-rigid distortions.

As shown in Fig. 3, assuming that there are infrared and visible image pairs $I_{ir}\in\mathbb{R}^{H \times W \times 1}$ and $I_{vi}\in\mathbb{R}^{H \times W \times 1}$, we use the pre-trained backbone network of MatchFormer\cite{wang2022matchformer} to extract multi-scale features for $I_{ir}$ and $I_{vi}$ to get the feature maps ${F}_{ir}^{1/k}\in \mathbb{R}^{\frac{H}{k} \times \frac{W}{k} \times C}$ and ${F}_{vi}^{1/k}\in \mathbb{R}^{\frac{H}{k} \times \frac{W}{k} \times C}$, where $k=\{4,8,16,32\}$ and $C$ corresponds to the number of channels for different feature maps. SAM (Semantic-aware module) senses and identifies the sROI in image pairs using the feature information $F_{ir}^{1/8}\in \mathbb{R}^{\frac{H}{8} \times \frac{W}{8} \times C}$ and $F_{vi}^{1/8}\in \mathbb{R}^{\frac{H}{8} \times \frac{W}{8} \times C}$ to get the sROI $R_{ir}\in \mathbb{R}^{\frac{H}{8} \times \frac{W}{8} \times 1}$ and $R_{vi}\in \mathbb{R}^{\frac{H}{8} \times \frac{W}{8} \times 1}$. The HDM (HOL-Deep hybrid matching) will perform feature matching based on sROI and uses the corresponding feature points to calculate TPS transformation to get the registration images. Finally, the registration infrared and visible images are fused using IFM (Image fusion module) to generate the fused image $I_{fusion}$.

In summary, the framework of SA-DNet consists of the following three steps:

1)  \emph{SAM perceives the sROI in infrared and visible images:} for the given infrared and visible image pairs, SAM perceive sROI in the images by mixing Local Window Self-Attention and Attention-Enhancing Strip Pool, described in Section III-A;

2)  \emph{HDM performs feature matching in sROI:} get better feature matching by combining Hierarchical Orientation Line (HOL) descriptors with deep feature and
search regional nearest neighbor matching in sROI, described in Section III-B;

3)  \emph{IFM performs fusion on the infrared and visible images:} fuse the registered images by combining feature map attention mechanism with Auto-Encoder architecture, described in Section III-C.

The total processing can be expressed in the cascade subtransform as:
\begin{equation}
(R_{ir},R_{vi})=SAM(F_{ir}, F_{vi})
\end{equation}
\begin{equation}
\begin{split}
 I_{fusion}=SFM(HDM((R_{ir},R_{vi}) \times (F_{ir}, F_{vi})) \\
            \stackrel{TPS}{\longrightarrow} (I_{ir},I_{vi}))
\end{split}
\end{equation}
\subsection{Semantic Awareness}
\textbf{Motivations.} In contrast to the dominating visual feature transformation in machine vision, human vision focuses their attention on task-relevant or specific semantic regions to reduce redundant processing of the remainder of the background\cite{rothkopf2007task}\cite{sullivan2012role}, thus facilitating efficient visual processing in the human brain. Based on this intuition, we choose to focus feature matching just on the task-relevant semantic regions and design SAM to enable SA-DNet to sense the semantic region of interest (sROI) in the images.

For infrared and visible image pairs $I_{ir}\in\mathbb{R}^{H \times W \times 1}$ and $I_{vi}\in\mathbb{R}^{H \times W \times 1}$, after obtaining the multi-scale feature map ${F}_{ir}^{1/k}$ and ${F}_{vi}^{1/k}$, SAM uses feature maps $F_{ir}^{1/8}\in \mathbb{R}^{\frac{H}{8} \times \frac{W}{8} \times C}$ and $F_{vi}^{1/8}\in \mathbb{R}^{\frac{H}{8} \times \frac{W}{8} \times C}$ to predict the sROI in infrared and visible images, which employs Local Window Self-Attention\cite{liu2021swin} and Attention-Enhancing Strip Pool to interact with the spatial feature information of the images. Note that the feature maps used in SAM are identical to those used by HDM, which means that the SAM only needs to provide HDM with prediction maps of 1/8 of the original image dimension.

\textbf{Local Window Self-Attention.} The emergence of Transformer\cite{vaswani2017attention} has made it a trend to capture contextual information by self-attention. Nevertheless, due to the high computational complexity, we have to make a trade-off between performance and computational complexity. Following\cite{liu2021swin}, we only use the Transformer encoder within a local window to increase perception of local semantic information, providing accurate edge information for sROI. 

Specifically, as illustrated in Fig. 4, we first divide the feature maps into different windows with the size of 8$\times$8, then restrict the attention computation to the non-overlapping local windows. The $\mathbf{Q}$ and $\mathbf{(K, V)}$ of the self-attention come from the same window. In calculating the self-attention, we use Scaled Dot-Product Attention\cite{vaswani2017attention}:
\begin{equation}
\operatorname{Attention}(Q, K, V)=\operatorname{softmax}\left(\frac{Q K^{T}}{\sqrt{d_{k}}}\right) V
\end{equation}
As the convolution has advantages in processing underlying features and visual structures, we use the convolution blocks after each Window Transformer block to further transform the feature maps and finally get the feature maps ${{F}}_{ir}^{window}$ and ${{F}}_{vi}^{window}$.

\textbf{Attention-Enhancing Strip Pool.} Since the self-attention is computed only inside a local window, interaction of remote spatial feature information is lost. To compensate for this part of feature information interaction and locate the spatial position information of sROI accurately, we develope Attention-Enhancing Strip Pool. It is modified based on strip pool\cite{hou2020strip}, which is an irregular spatial pool that averages the features in each row and column of the feature map to obtain $\mathbf{y}^{h}\in\mathbb{R}^{H \times 1 \times C}$ and $\mathbf{y}^{w} \in \mathbb{R}^{1 \times W \times C}$:
\begin{equation}
y_{i}^{h}=\frac{1}{W} \sum_{0 \leq j<W} x_{i, j}
\end{equation}
\begin{equation}
y_{j}^{w}=\frac{1}{H} \sum_{0 \leq i<H} x_{i, j}
\end{equation}
Due to the long and narrow shape of the kernel, the receptive field of strip feature maps $\mathbf{y}^{h}$ and $\mathbf{y}^{w}$ are only the rows and columns of the feature maps. As a sequence-shaped feature map, it is advantageous to utilize self-attention and cross- attention for strip feature map to improve the receptive field. After use self-attention for $\mathbf{y}^{h}$ and $\mathbf{y}^{w}$, we obtain the transformed strip feature maps $\tilde{\mathbf{y}}_{}^{h}\in \mathbb{R}^{H \times 1 \times C}$ and $\tilde{\mathbf{y}}_{}^{w}\in \mathbb{R}^{1 \times W \times C}$, interpolate $\tilde{\mathbf{y}}_{}^{h}$ and $\tilde{\mathbf{y}}_{}^{h}$ to get feature maps $\tilde{\mathbf{y}}_{i}^{h}\in \mathbb{R}^{H \times W \times C}$ and $\tilde{\mathbf{y}}_{i}^{w}\in \mathbb{R}^{H \times W \times C}$. Finally, $\tilde{\mathbf{y}}_{i}^{h}$ and $\tilde{\mathbf{y}}_{i}^{w}$ are combined to get the feature map ${{F}}_{ir}^{strip}$ and ${{F}}_{ir}^{strip}$ by:
\begin{equation}
\tilde{y}_{i, j, c}=\tilde{y}_{i, c}^{h}+\tilde{y}_{j, c}^{w}
\end{equation}
\begin{figure*}[htbp]
\centering
\includegraphics[width=180mm]{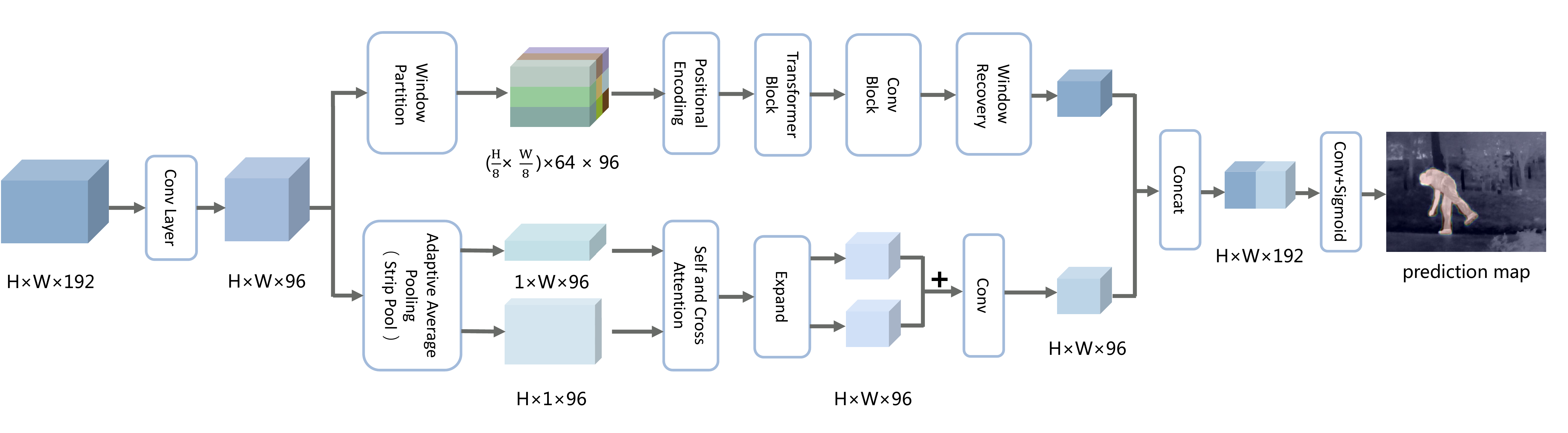}
\caption{\textbf{The specific network structure of SAM.} The feature maps $F_{ir}^{1/8}$ and $F_{vi}^{1/8}$ obtained through the backbone network first reduce the channel dimension by convolution blocks. Then obtain the feature maps ${{F}}_{}^{window}$ and ${{F}}_{}^{strip}$ by Local Window Self-Attention and Attention-Enhancing Strip Pool. The ${{F}}_{}^{window}$ and ${{F}}_{}^{strip}$ are concatenated and fused through convolution block, finally outputs the prediction map ${{R}}^{ir}$ and ${{R}}^{vi}$ using the sigmoid function.}
\label{fig2:env}
\end{figure*}
\textbf{Semantic Region Prediction.} After obtaining the local and global features in the images using Local Window Self-Attention and Attention-Enhancing Strip Pool, we concatenation ${{F}}_{}^{window}$ and ${{F}}_{}^{strip}$, then integrate the feature information by multiple convolution blocks. Finally, the Sigmoid activation function is used to output the sROI prediction map ${R}^{ir}$ and ${R}^{vi}$, which can be expressed as:
\begin{equation}
{{R}}^{ir}=Sigmoid(Conv(Concat({{F}}_{ir}^{window},{{F}}_{ir}^{strip})))
\end{equation}
\begin{equation}
{{R}}^{vi}=Sigmoid(Conv(Concat({{F}}_{vi}^{window},{{F}}_{vi}^{strip})))
\end{equation}


\subsection{Regional Feature Matching}
\textbf{Motivations.} MatchFormer\cite{wang2022matchformer}, LoFTR\cite{sun2021loftr}, and other methods\cite{giang2022topicfm}\cite{sarlin2020superglue} can rely on their powerful matching ability to get many corresponding feature points in the natural scenes. In the feature mathcing process, MatchFormer progressively differentiate the features at different spatial position of the images by self-attention and cross-attention. However, the small area of the sROI and the modal differences between infrared and visible images render these features insufficient for reliable discrimination in all scenes, thus bringing negative effects on the subsequent feature matching steps. For this reason, we develop Hierarchical Orientation Line (HOL) Descriptors and integrate them with deep features to identify the most possible matches inside sROI based on their matching score.


\textbf{HOL Descriptors.} As shown in Fig. 5, HOL describes the distribution of feature point positions inside sROI. Same as the dense feature matching method, the distribution of feature points in SA-DNet is uniform, and these feature points are matched by constructing robust feature descriptors. Assume that the feature point sets $\mathcal{N}_{ir}\in \mathbb{R}^{M \times 2}$ and $\mathcal{N}_{vi}\in \mathbb{R}^{N \times 2}$ in sROI of infrared and visible images, respectively. Each pair of feature points in $\mathcal{N}_{ir}$ and $\mathcal{N}_{vi}$, denoted by $x_{ir}=\left(x_{i}, y_{i}\right)$ and $x_{vi}=\left(x_{j}, y_{j}\right)$, can be described in terms of the number of feature points surrounding $x_{ir}$ and $x_{vi}$ in the four cardinal directions: top, bottom, left and right. In the HOL descriptor, we give a concentric square expansion of the neighborhoods for $x_{ir}$ and $x_{vi}$ to gradually count the histograms of the different layers. The neighborhood feature points of each layer can be calculated as:
\begin{equation}
\mathcal{N}_{ir}^{k}=\left\{x_{ir}^{k} \mid x_{ir}^{k} \in \mathcal{N}_{ir}, d\left(x_{ir}^{k}, x_{ir}\right)=k\right\}
\end{equation}
\begin{equation}
\mathcal{N}_{vi}^{k}=\left\{x_{vi}^{k} \mid x_{vi}^{k} \in \mathcal{N}_{vi}, d\left(x_{vi}^{k}, x_{vi}\right)=k\right\}
\end{equation}
where $d(\cdot)$ denotes the Chebyshev distance function between two points. As a method of dense feature matching, the Chebyshev distance between any two feature points is a multiple of 8 and we take ${k}=\{k_i \mid k_i \in(0, 240),k_i\bmod8=0\}$. 

\begin{figure}[htbp]
\centering
\includegraphics[width=90mm]{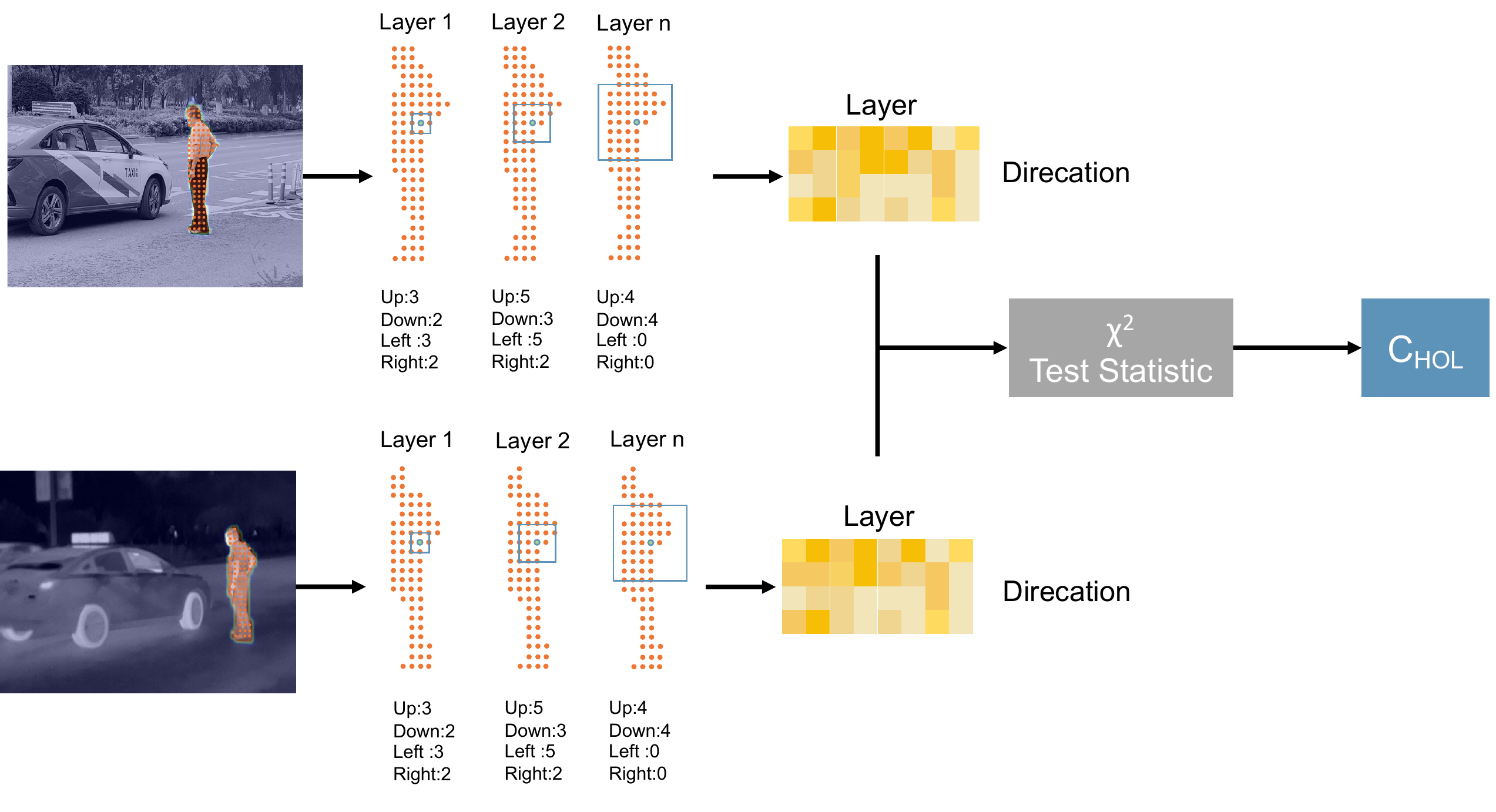}
\caption{\textbf{HOL descriptor.} HOL describes the feature points distribution information in sROI. For a given feature point, we continuously expand the direction line to perceive the feature points distribution at different layers. After get the histogram containing the feature points distribution at different levels in four directions, the cost matrix is calculated based on the histogram of all feature points using $\chi^2$ test statistic.}
\label{fig2:env}
\end{figure}
After computing the histogram for the number of feature points in different directions from different layers, we can obtain the HOL descriptors ${D}_{ir}\in \mathbb{R}^{N \times 116}$ and ${D}_{vi}\in \mathbb{R}^{N \times 116}$ of $\mathcal{N}_{ir}\in \mathbb{R}^{N \times 2}$ and $\mathcal{N}_{vi}\in \mathbb{R}^{N \times 2}$. Similarly to some descriptors constructed based on statistics, we use $\chi^2$ test statistic to compute the cost matrix for matching feature point descriptors in sROI.
\begin{equation}
C_{HOL}\left(x_{ir}, x_{vi}\right)=\frac{1}{2} \sum_{i \in k} \frac{\left[{h}_{ir}^{i}-{h}_{vi}^{i}\right]^2}{{h}_{ir}^{i}+{h}_{vi}^{i}}
\end{equation}
where $C_{HOL}\left(x_{i r}, x_{v i}\right)$ denotes the matching cost of feature points $x_{ir}$ and $x_{ir}$, ${h}_{ir}$ and ${h}_{vi}$ are the HOL descriptors of $x_{ir}$ and $x_{ir}$, ${h}_{ir}^{i}$ and ${h}_{vi}^{i}$ denote the statistical histograms for each layer in ${h}_{ir}$ and ${h}_{vi}$.

\textbf{HOL-Deep feature matching.} If the edge information of sROI is precisely enough, we can find the corresponding feature points inside sROI only based on $C_{HOL}\left(x_{i r}, x_{v i}\right)$. To cope with the case that sROI may be inaccurate and achieve a more accurate feature matching, we combine the matching information given by deep features $F_{ir}^{1/8}\in \mathbb{R}^{\frac{H}{8} \times \frac{W}{8} \times C}$ and $F_{vi}^{1/8}\in \mathbb{R}^{\frac{H}{8} \times \frac{W}{8} \times C}$ with $C_{HOL}\left(x_{i r}, x_{v i}\right)$ to find the optimum corresponding feature points. Following the coarse matching procedure of LoFTR\cite{sun2021loftr} and MatchFormer\cite{wang2022matchformer}, the deep feature descriptors $F_{ir}^{1/8}\in \mathbb{R}^{\frac{H}{8} \times \frac{W}{8} \times C}$ and $F_{vi}^{1/8}\in \mathbb{R}^{\frac{H}{8} \times \frac{W}{8} \times C}$ are used to calculate the score matrix of the feature points by following equation: 
\begin{equation}
\mathcal{S}_{Deep}(x_{ir}, x_{vi})= \left\langle{F}_{ir}^{1/8}, {F}_{vi}^{1/8}\right\rangle(x_{ir},x_{vi})
\end{equation}

After obtaining $C_{HOL}\left(x_{i r}, x_{v i}\right)$ and $S_{Deep}\left(x_{i r}, x_{v i}\right)$, it is worth considering how to combine them to obtain the optimal assignment matrix. Based on the following two reasons, we cannot simply add the $C_{HOL}\left(x_{i r}, x_{v i}\right)$ and $S_{Deep}\left(x_{i r}, x_{v i}\right)$ in a weighted way.

1) \emph{Feature points closer to the sROI edge are more sensitive to edge error.} In HOL descriptors, the shape distribution of feature points at the edge of the sROI provides the relative position information of each feature point. For feature points inside the edges, the spatial position distribution is uniform, resulting in identical local structure information for each feature point. However, if we construct HOL descriptors for feature matching using only sROI edge feature points, the matching performance will be significantly reduced. This is because the feature points are closer to the sROI edge, the more sensitive HOL descriptors are to the edge error variation, which makes a significant difference discrepancy ${h}_{ir}$ and ${h}_{vi}$. In contrast, when the feature points are further from the sROI edge, they are more robust to edge errors. 

2) \emph{Inaccurate sROI will cause large differences between sROI area in infrared and visible images.} When capturing infrared and visible images, we usually adjust the different imaging devices to the same focal length to facilitate the process of image registration, which means that the area of sROI in ${I}_{ir}$ and ${I}_{vi}$ will be similar. Based on this assumption, we calculate the ratio $\sigma$ $(0<\sigma<1)$ of sROI area in ${I}_{ir}$ and ${I}_{vi}$, the smaller $\sigma$ means that the difference between sROI area is larger and the edge of sROI is more likely to be inaccurate.

From the above observations, we design a Gaussian-weighted decay strategy by comparing the difference of sROI area in ${I}_{ir}$ and ${I}_{vi}$ and the distance of the feature point to the center of mass in each sROI.

Specifically, the matching score matrix that combines the deep features and HOL features can be expressed as:
\begin{equation}
\mathcal{S}(x_{ir}, x_{vi})= \mathcal{S}_{deep}(x_{ir}, x_{vi})-\delta\cdot\sigma\cdot\lambda\cdot C\left(x_{ir}, x_{vi}\right)
\end{equation}
\begin{equation}
\lambda=\frac{1}{\sigma \sqrt{2 \pi}} (e^{-\frac{(\frac{ d(x_{ir},x_{ir}^c)}{d_{ir}^{max}})^2}{2 \sigma^2}}+ e^{-\frac{(\frac{ d(x_{vi},x_{vi}^c)}{d_{vi}^{max}})^2}{2 \sigma^2}})
\end{equation}

\begin{equation}
\sigma=\omega\cdot\frac{\min \left(S\left(R_{ir}\right), S\left(R_{vi}\right)\right)}{\max \left(S\left(R_{ir}\right), S\left(R_{vi}\right)\right)} ,\sigma \in(0,1], \omega \in(0,1]
\end{equation}
where $\delta$ denotes the weight parameter, $\omega$ is a hyperparameter, $x_{ir}^c$ and $x_{vi}^c$ denote the center of mass in sROI, $d_{ir}^{max}$ and $d_{vi}^{max}$ denote the farthest distance from $x_{ir}^c$ and $x_{vi}^c$ to the sROI edge. $S(\cdot)$ denotes the area of sROI, in the actual calculation, we set the number of feature points inside the sROI as its area. $\lambda$ is a Gaussian function that represents the weighting factor of the HOL when combining with the deep feature, a larger $\lambda$ means that the HOL descriptor has large influence on the final feature matching result. The value of $\lambda$ is influenced by two factors. On the one hand, when the difference between the sROI areas of infrared and visible images is large, the sROI may be inaccurate, causing $\sigma$ to take a smaller value, and the value of $\omega$ will increase, which reduces the influence of HOL on the feature matching results. On the other hand, when the feature point is closer to the edge of sROI, HOL descriptor will be sensitive to the edge error, the value of $\omega$ become increasing, which also reduces the influence of HOL on the feature matching results.

Based on $\mathcal{S}(x_{ir}, x_{vi})$, we search for feature points corresponding to the maximum score in the score matrix of each feature point position in sROI between infrared and visible images. In addition, we do not search for corresponding points outside the sROI nor treat feature points outside the sROI as potential corresponding feature points. If the corresponding score of two points in the score matrix is greater than a certain threshold $\theta$, the feature points are considered as corresponding feature points. Do the same operation for each feature point inside the sROI to get the corresponding feature set $\mathcal{C}$. In this way, we use sROI to guide feature matching and reduce the mismatched feature points while increasing the number of corresponding feature points.
\begin{equation}
\mathcal{C}=\left\{(i, j) \mid \forall i \in R^{ir}, \forall j \in R^{vi},\mathcal{S}(i, j) \geq \theta\right\}
\end{equation}

After getting the corresponding feature points set $\mathcal{C}$, we using RANSAC\cite{fischler1981random} for mismatch removal and leverage the remaining feature points to estimate the TPS transformation parameters. Finally, the well-registered sROI in the images can be used in the subsequent visual tasks.     

\subsection{Image Fusion}
SA-DNet focuses on providing a preliminary solution for semantic-aware on-demand registration. To visually demonstrate its advantages, we simplify the implementation of image fusion and use Auto-Encoder based structure to fuse infrared and visible images. IFM consists of three steps: Encoder, Feature Fusion and Decoder. The Encoder uses ${F}_{ir}^{1/4}\in \mathbb{R}^{\frac{H}{4} \times \frac{W}{4} \times 128}$ and ${F}_{vi}^{1/4}\in \mathbb{R}^{\frac{H}{4} \times \frac{W}{4} \times 128}$ to generate spatial attention maps ${F}_{ir}^{a}\in \mathbb{R}^{H \times W \times 1}$ and ${F}_{ir}^{a}\in \mathbb{R}^{H \times W \times 1}$ by convolutional block and spatial attention mechanism to guide the feature fusion. 
\begin{equation}
{F}_{ir}^{a}=Conv1(UP({F}_{ir}^{1/4}))\times Conv2(UP({F}_{ir}^{1/4}))
\end{equation}
\begin{equation}
{F}_{vi}^{a}=Conv1(UP({F}_{vi}^{1/4}))\times Conv2(UP({F}_{vi}^{1/4}))
\end{equation}
where $UP(\cdot)$ denotes the upsampling operation, $Conv1$ and $Conv2$ denote the different convolution blocks.

Since the backbone network lacks sufficient shallow features, in order to obtain better quality fused images, we add several convolution blocks to extract shallow features from the original infrared and visible images. The shallow feature maps are multiplied with the attention maps to get ${F}_{ir}^{s}\in \mathbb{R}^{H \times W \times 48}$ and ${F}_{vi}^{s}\in \mathbb{R}^{H \times W \times 48}$. Then ${F}_{ir}^{s}$ and ${F}_{vi}^{s}$ will be fused by SCA in\cite{li2020nestfuse}, and the fused features are passed through the Decoder consisting of multiple convolutional layers to output the fused image.

\subsection{Loss Function}
SA-DNet is trained in two stages, the first stage uses a method similar to visual saliency detection for training the SAM, the data with mask annotations allows SAM to learn sROI in infrared and visible images. Assume the predicted map ${R}$ and the labeled map ${Y}$, a weighted BCE loss function is applied to train SA-DNet:
\begin{equation}
L_{SAM}=-\frac{1}{N}\sum_{i=1}^{N}[w\cdot y_{i} \cdot \log r_{i}+ (1-w)\cdot (1-y_{i}) \cdot \log (1-r_{i})]
\end{equation}
where $y_{i}$ denotes the pixel in ${Y}$, which the value is 0 or 1 and indicate whether the point is in sROI. ${N}$ is the area of ${R}$ and ${w}$ is the hyperparameter, indicating the weight of the loss.

The second stage requires training of the IFM in the form of image reconstruction. Suppose the input image for the image reconstruction process is ${I}_{in}$ and the output image is ${I}_{out}$. We train the Encoder and Decoder using the SSIM loss function and ${L1}$ loss function.

Structural similarity (SSIM)\cite{wang2004image} is widely used in image fusion methods to preserve texture detail information in visible images and is expressed as:
\begin{equation}
L_{SSIM}=1-S S I M\left(I_{out}, I_{in}\right)
\end{equation}

The ${L1}$ loss function enable the brightness and color in the reconstructed images to be kept constant for compensating the deficiency of the SSIM loss function:
\begin{equation}
L_1=\frac{1}{H \times W} \sum\|I_{out}-I_{in}\|
\end{equation}

After get $L_{\text {SSIM }}$ and $L_{\text {1}}$, the overall loss of IFM is calculated as:
\begin{equation}
L_{IFM}=L_{\text {SSIM }} \times 0.3+L_\text{1}\times 0.7
\end{equation}

\section{Experiments}
In this section we first outline the implementation details and the IVS dataset we built for training, followed by extensive experiments to illustrate the rationality and superiority of SA-DNet:
1) \emph{Ablation Studies.} Illustrating the rationality and effectiveness of the components in SA-DNet through four ablation studies experiments.
2) \emph{Semantic Region Feature Matching.} By comparing the performance of SA-DNet and other five advanced feature matching methods in semantic region feature matching to demonstrate the superiority of SA-DNet when feature matching inside sROI.
3) \emph{Applications.} To demonstrate the scalability of SA-DNet, we conduct experiments including Image Fusion, Object Detection and Structure from Motion (SfM).


\textbf{Details of Implementation.}
In this paper, we propose the concept of Semantic-Aware on-Demand registration. Nonetheless, there is no publicly available dataset for different semantic object segmentation in infrared and visible images. To facilitate the building of datasets and experiments, we limit the specific semantic aware object to the pedestrian and build IVS dataset for pedestrian to train SA-DNet.
In the first training phase, we employed the backbone in MatchFormer\cite{wang2022matchformer} as a pre-trained model, and the backbone parameters were fixed throughout the training process. SAM was trained using the IVS dataset and the Adam optimizer, minimizing the loss $L_{\text {SAM}}$, with a learning rate of 1e-3 and the parameter ${w=0.7}$. The batch size is 4, and stopping training after 50 epochs.
In the second training phase, we randomly selected 20,000 images from the COCO\cite{lin2014microsoft} dataset for training IFM, using the Adam optimizer, minimizing the loss $L_{\text {IFM}}$, with a learning rate of 1e-4. The batch size is 4, and stopping training after 5 epochs. All training and experiments were conducted on a PC with a 2.90GHz Intel Core i5 10400 CPU and NVIDIA GeForce GTX 1660.

\textbf{IVS Dataset.}
We found that there was no publicly available dataset of infrared image with mask annotation for semantic object. Therefore, we build the IVS dataset, which contains 4691 infrared images and the corresponding pedestrians mask. The mask of each image was obtained by SegFormer\cite{xie2021segformer}, and the incorrectly segmented portions of the images were eliminated manually.
\begin{table*}[!t]
\renewcommand\arraystretch{1.5}
        \centering
        \caption {Compare the effect of local region feature matching with Gaussian-weighted decay strategy and different $\omega$ values.}
        \label{tab:chap:table_1}
        \begin{tabular}[c]{cccc}
                \toprule
                \textbf{Method\textbackslash Metrics} & \textbf{Matches} & \textbf{Correct Matches} & \textbf{Matches Accuracy} \\
                \midrule
                w/o Gaussian-weighted decay strategy &75.8 &46.2 & 0.609 \\
                \midrule
                $\omega$=0.3 &72.9 &45.4 &0.622 \\  
                $\omega$=0.4 &77.1 &49.0  &0.636 \\    
                $\omega$=0.5 &80.3 &51.4  &\textbf{0.640} \\  
                $\omega$=0.6 &81.7 &\textbf{52.2}  &0.638 \\    
                $\omega$=0.7 &\textbf{82.2} &\textbf{52.2}  &0.635 \\  
                $\omega$=0.8 &\textbf{82.2} &52.0  &0.633 \\  
                $\omega$=0.9 &81.7 &51.4  &0.628 \\  
                $\omega$=1.0 &80.9 &51.0  &0.630 \\  
                \bottomrule
        \end{tabular}
\end{table*}
\subsection{Ablation Studies}
To evaluate the improvement of performance that different components can provide, we perform four ablation studies.

\textbf{(i)} Use only Local Window Self-Attention (LWS) or Attention-Enhancing Strip Pool (AESP) in SAM. As can be seen from Fig. 6, when using LWS or AESP alone, the ability to perceive pedestrians is reduced, resulting in the presence of a large amount of noise that fails to provide a satisfactory semantic object localization and fine edge segmentation. In fact, hybrid attention is more likely to be the ideal option than single attention. When LWS is combined with AESP, SAM is better able to preserve the local and global feature information to integrate the contextual semantic information in the images, leading to accurate perception of semantic object in the image based on the need of different task.

\begin{figure}[htbp]
\centering
\includegraphics[width=90mm]{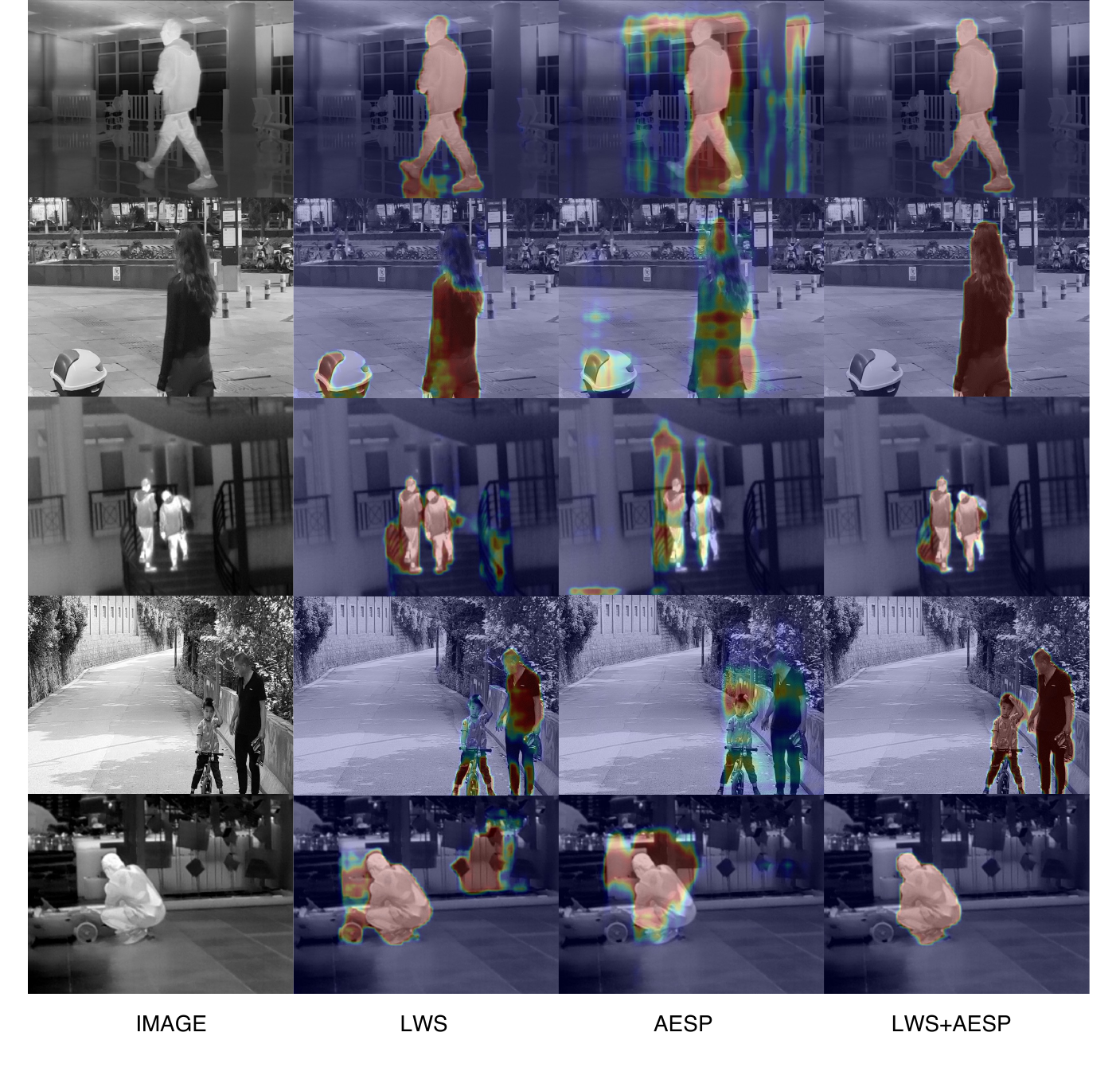}
\caption{Ablation study of mixed attention by Local Window Self-Attention (LWS) and Attention-Enhancing Strip Pool (AESP).}
\label{fig2:env}
\end{figure}
\begin{figure*}[htbp]
\centering
\includegraphics[width=180mm]{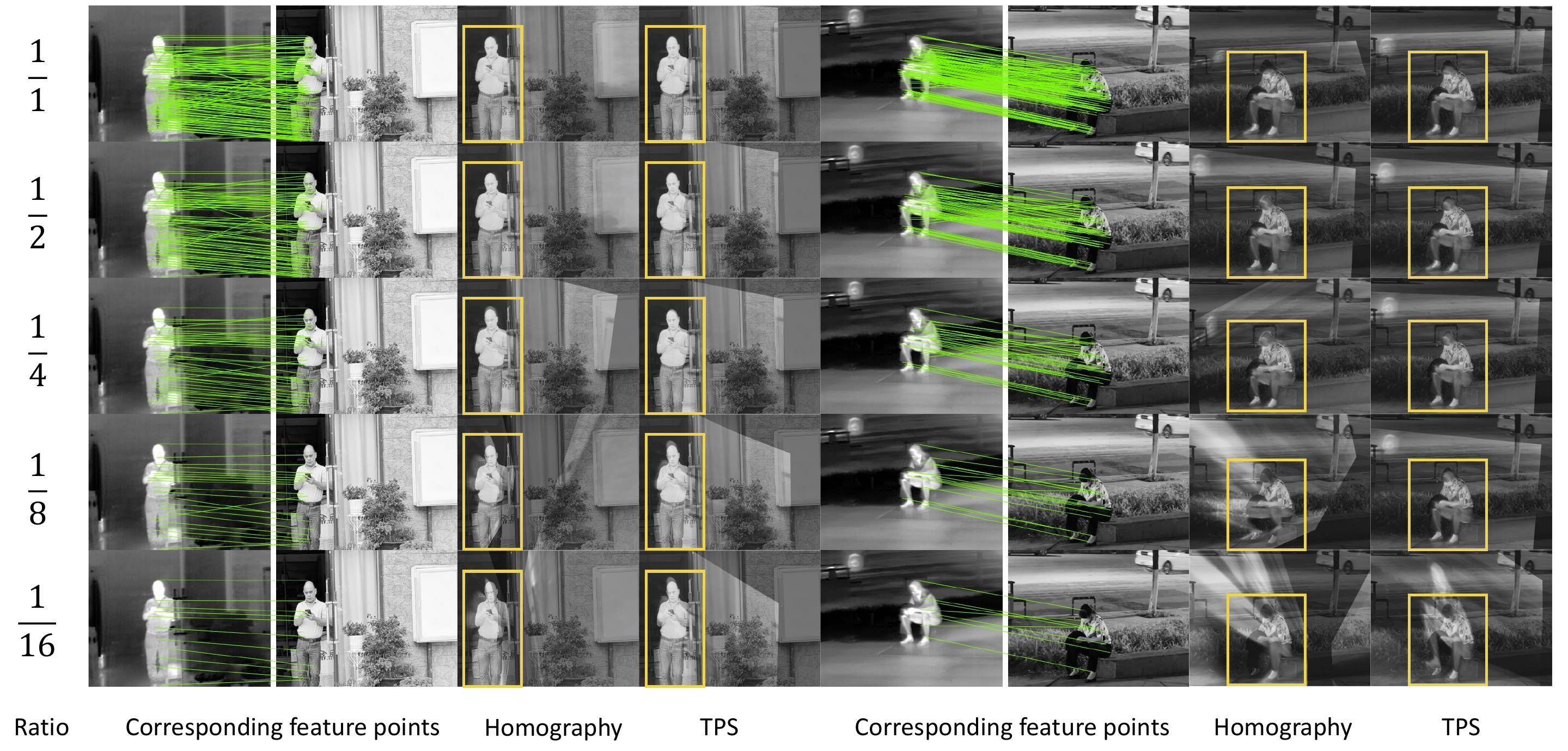}
\caption{Comparison of the robustness of Homography and TPS when retaining different ratios of the corresponding feature points. It can be observed that both Homography and TPS can perform accurate registration and fusion of semantic objects when a sufficient number of corresponding feature points are retained. As the number of corresponding feature points decreases, the accuracy of the Registered images transformed using Homography becomes worse, which leads to bad visual effects. In contrast, TPS still has better robustness with fewer matched feature points.}
\label{fig2:env}
\end{figure*}
\begin{figure*}[htbp]
\centering
\includegraphics[width=180mm]{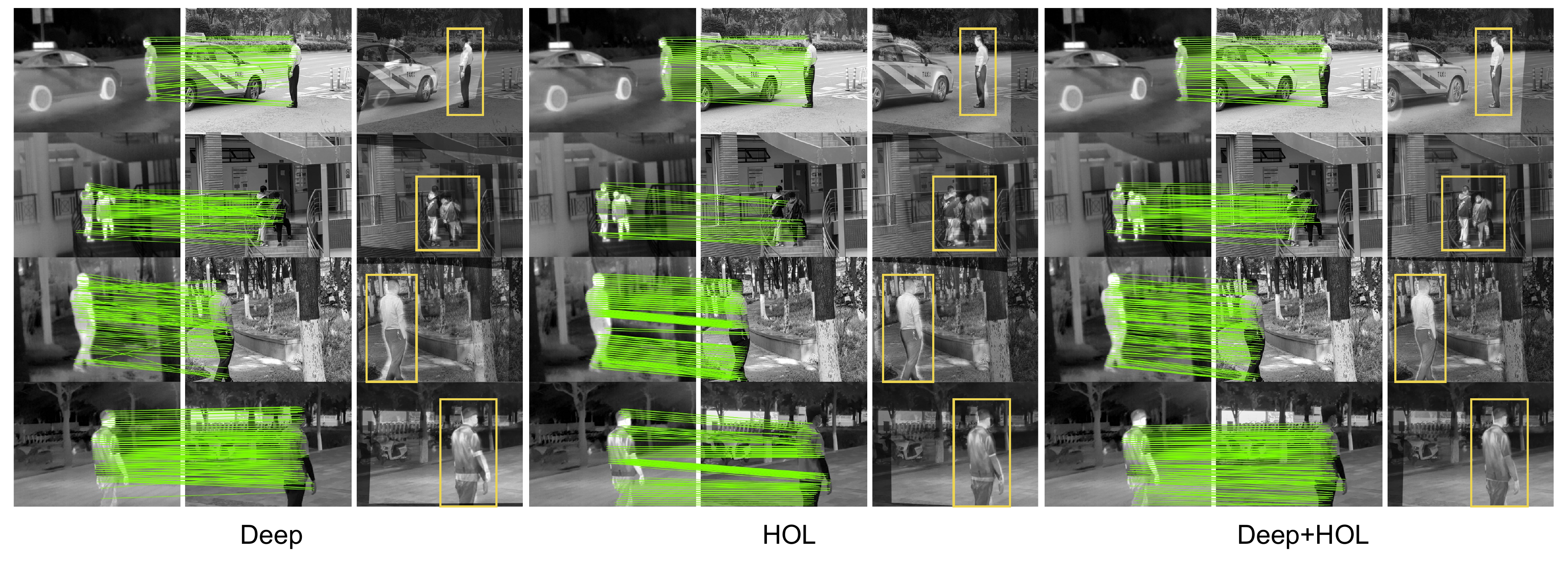}
\caption{\textbf{Ablation study of feature matching using different feature descriptors.} It shows that the quality of image registration is better when combining Deep and HOL features than when using Deep or HOL features alone.}
\label{fig2:env}
\end{figure*}

\textbf{(ii)} Replace TPS transformation with Homography transformation. Homography is often used as a transformation model for various image registration due to its simplicity. Assuming that corresponding feature points $X_{i}=\left(x_{i}, y_{i}\right)$ and $X_{i}^{\prime}=\left(x_{i}^{\prime}, y_{i}^{\prime}\right)$ are obtained from infrared and visible semantic regions, the Homography can be defined as:
\begin{equation}
\left(\begin{array}{c}
x_{i}^{\prime} \\
y_{i}^{\prime} \\
1
\end{array}\right)=H\left(\begin{array}{c}
x_{i} \\
y_{i} \\
1
\end{array}\right)
\end{equation}
where H represents the Homography matrix. 
In contrast to Homography, TPS obtains the transformation coefficients by solving the linear system:
\begin{equation}
\theta=\left(\begin{array}{cc}
\mathcal{K} & X^{\prime} \\
X^{\prime T} & \mathbf{O}_{3 \times 3}
\end{array}\right)^{-1}\left(\begin{array}{c}
X \\
\mathbf{O}_{3 \times 2}
\end{array}\right)
\end{equation}
where $\mathcal{K}_{n \times n}$ is a radial basis kernel. By solving $\theta$, we sample pixels from the infrared image to obtain the transformed image, and other points on the image can be rectified by interpolation.

\begin{table}[!t]
\renewcommand\arraystretch{1.5}
        \centering
        \caption {Comparison of six methods for feature matching in local semantic regions.}
        \label{tab:chap:table_1}
        \begin{tabular}[c]{cccc}
                \toprule
                \textbf{Method\textbackslash Metrics} & \textbf{Matches} & \textbf{Correct Matches} & \textbf{Matches Accuracy} \\
                \midrule
                RIFT & 18.7 & 14.5  & \textbf{0.775} \\
                R2D2 & 0.9 & 0.4  & 0.444 \\  
                SP+SG & 16.9 & 11.4  & 0.675 \\    
                LoFTR & 26.7 & 10.5  & 0.393 \\  
                MatchFormer & 58.4 & 35.6  & 0.610 \\    
                \textbf{SA-DNet} & \textbf{105.9} & \textbf{72.4}  & 0.684 \\  
                \bottomrule
        \end{tabular}
\end{table}

\begin{table*}[ht]
\renewcommand\arraystretch{1.5}
        \centering
        \caption {Quantitative comparison of registration and fusion effects of different feature matching methods.}
        \label{tab:chap:table_1}
        \begin{tabular}[c]{m{2.5cm}<{\centering}m{1.7cm}<{\centering}m{1.7cm}<{\centering}m{1.7cm}<{\centering}m{1.7cm}<{\centering}m{1.7cm}<{\centering}m{1.7cm}<{\centering}}  
                \toprule
                \textbf{Method\textbackslash Metrics} & \textbf{AG $\uparrow$} & \textbf{CE $\downarrow$} & \textbf{EI $\uparrow$}  &\textbf{MI $\uparrow$} &\textbf{SSIM $\uparrow$} &\textbf{CT $\uparrow$}\\ 
                \midrule
                 RIFT & 6.4296 & 1.0377 &  \textbf{66.5973}   & 0.7739  & 0.1432 & 168.6796\\
               R2D2 & 6.0890 & 1.4427  & 62.0954 & 0.4110  & 0.1086 & 161.9549\\  
  SP+SG & 6.4017 & 1.1597  & 65.3272 & 0.6835  & \textbf{0.1517} & 167.6469\\    
  LoFTR & 6.0369 & 1.1447 & 61.5027  & 0.5959  & 0.1157 & 160.4468\\  
  MatchFormer & 6.2607 & 1.0796  & 64.0253  & 0.6543  & 0.1338 & 158.9639\\    
  \textbf{SA-DNet} & \textbf{6.4342} & \textbf{1.0335}  & 65.8174  & \textbf{0.8008}  &  0.1487 & \textbf{169.4297}\\  
                \bottomrule
        \end{tabular}
\end{table*}
After getting the corresponding feature points in sROI, we gradually reduce the number of corresponding feature points to evaluate the robustness of Homography and TPS. The registration and fusion results are shown in Fig. 7, It can be found that when we get enough corresponding feature points inside sROI, the registration performance of Homography and TPS is similar. However, as the number of corresponding feature points decreases, the robustness of Homography becomes worse gradually relative to TPS. Based on this observation, SA-DNet adopt TPS as the transformation model to make better adaptation to different non-rigid transformations between the images.

\textbf{(iii)} Using Deep or HOL feature alone. As shown in Fig. 8, the modal differences between infrared and visible images result in reduced differentiation of deep features, which rendering the visual results of fused images unsatisfactory in some cases when matching using only the deep feature. HOL only utilizes the structural information of sROI without considering other potentially useful feature information, which making it unable to achieve accurate registration in some circumstances. In comparison, selecting the corresponding feature points by Deep+HOL features is a more reasonable way, because combining two complementary feature information can get more robust corresponding feature points, thus improving the performance of image registration.

\textbf{(iv)} Ablation study of Gaussian-weighted decay strategy and hyperparameters $\omega$ in Equation (15). We select 10 pairs of images in the M3FD\cite{liu2022target} dataset in which our SAM failed to sense sROI accurately to evaluate the effect of Gaussian-weighted decay strategy when coping with inaccurate sROI. Tab. I shows that the Gaussian-weighted decay strategy can improve the effect of feature matching in the case of inaccurate sROI. In addition, the Gaussian-weighted decay strategy can achieve the best accuracy when $\omega$=0.5, which proves that the Gaussian-weighted decay strategy in HDM is reasonable.

\begin{figure*}[htbp]
\centering
\includegraphics[width=180mm]{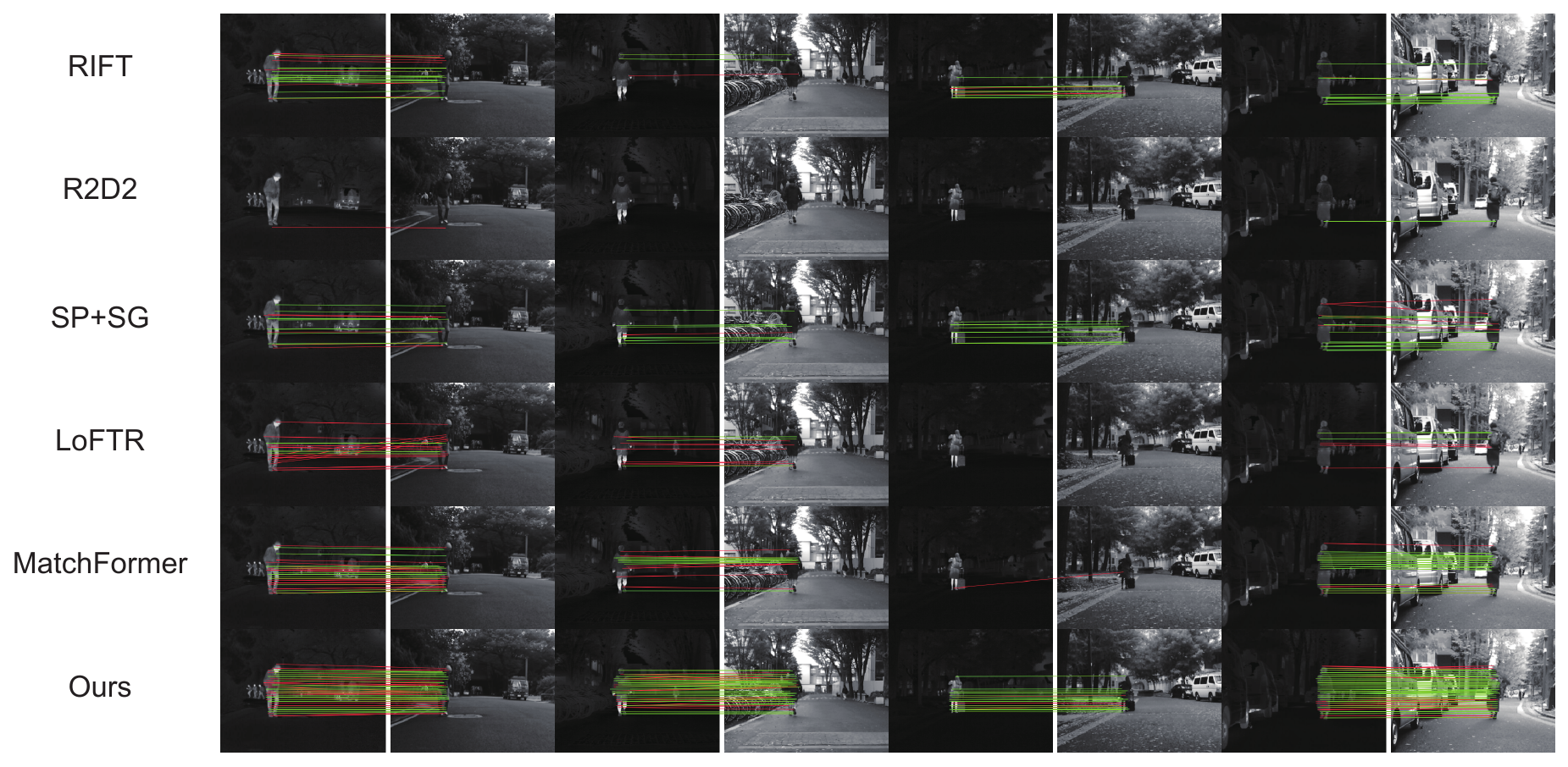}
\caption{\textbf{Comparison of SA-DNet with other five methods for feature matching within semantic objects.} The inliers (green lines) and outliers (red lines) measuring the performance of feature matching. SA-DNet achieves more corresponding feature points in various scenarios and maintains a high matching accuracy.}
\label{fig2:env}
\end{figure*}

\subsection{Results on Semantic Region Feature Matching}
To visualize the feature matching performance of SA-DNet inside sROI, we compare it with five feature matching methods: RIFT\cite{li2019rift}, R2D2\cite{revaud2019r2d2}, SuperPoint\cite{detone2018superpoint}+SuperGlue\cite{sarlin2020superglue}, LoFTR\cite{sun2021loftr}, and MatchFormer\cite{wang2022matchformer}. 
RIFT\cite{li2019rift} uses phase congruency (PC) for feature point detection, and uses maximum index map for feature description, and achieves good results for feature matching of multimodal images.
R2D2 proposes the concept of descriptor reliability, which masks the descriptors of some unreliable regions and gives more reliable feature matching.
SuperPoint\cite{detone2018superpoint}+SuperGlue\cite{sarlin2020superglue} is one of the well-performing detector-based methods, with advanced levels of matching accuracy and matching efficiency.
LoFTR\cite{sun2021loftr} introduces Transformer\cite{vaswani2017attention} into feature matching, which improves the matching effect of detector-free based methods to an unprecedented level. 
MatchFormer\cite{wang2022matchformer} is modified based on LoFTR to extract more efficient features in the backbone network using Transformer encoder, and gets better feature matching results.

All comparison methods are based on publicly available codes and use a lower matching threshold to obtain as many corresponding feature points as possible, where SuperPoint+SuperGlue, LoFTR, MatchFormer adjust the default matching threshold from 0.2 to 0.05.
We selected 38 pairs of infrared and visible images with pedestrian objects from the MSRS\cite{Tang2022PIAFusion} dataset. These images are registered, so we can easily find out which points are mismatched in the corresponding feature points. After obtaining the corresponding feature points by different methods, we only keep the feature points inside sROI and reject the feature points that are not inside sROI. The evaluation metrics include mean number of matches, mean number of correct matches, and mean match accuracy. We calculate the Euclidean Distance between feature points and their true correspondences, then treat the feature point pair with a distance greater than 8 as mismatched feature point.

The experimental results shown in Fig. 9 and Table II indicates that SA-Det has a good robustness when feature matching is performed in semantic regions, which obtain more reliable corresponding feature points in various scenes while maintaining favorable matching accuracy. The satisfactory matching performance provides a prerequisite for various visual tasks. On the other hand, the modal differences between infrared and visible images lead to LoFTR and R2D2 failing to get the corresponding feature points from sROI in some scenes. As a method designed specifically for multimodal image feature matching, RIFT achieves a high matching accuracy, but a fewer number of reliable corresponding feature points may not enable the TPS transformation to obtain precise parameter estimates.

\begin{figure*}[htbp]
        \centering
        \subfloat{
        \begin{minipage}{0.31\linewidth}
        \begin{center}
        \footnotesize{\textsf{RIFT}\cite{li2019rift}}
        \end{center}    
                \centering
                \includegraphics[width=1\linewidth]{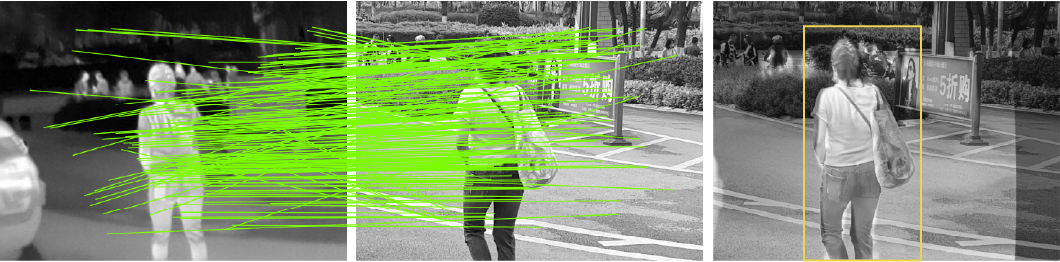}
                \includegraphics[width=1\linewidth]{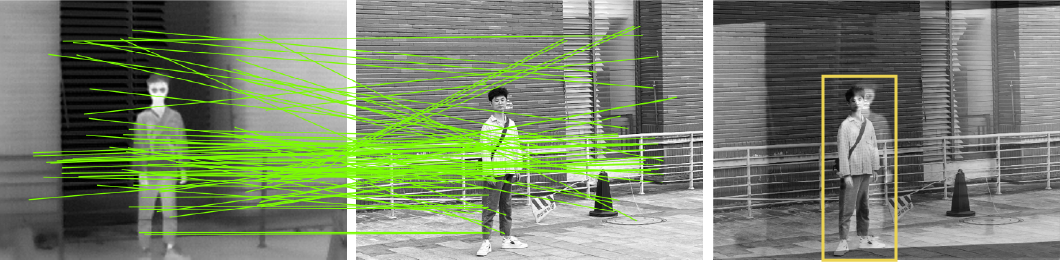}
                \includegraphics[width=1\linewidth]{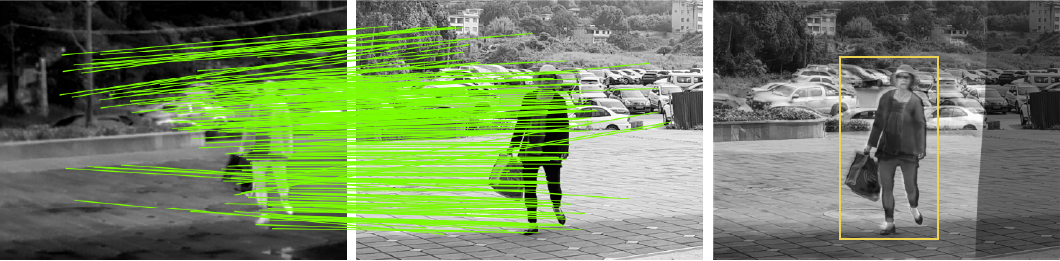}
                \includegraphics[width=1\linewidth]{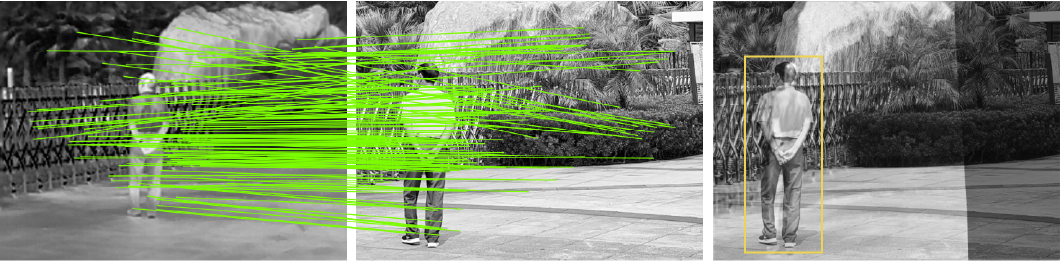}
                \includegraphics[width=1\linewidth]{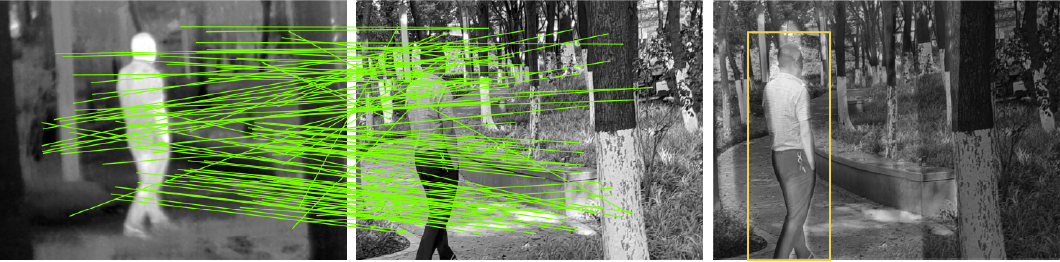}
                \includegraphics[width=1\linewidth]{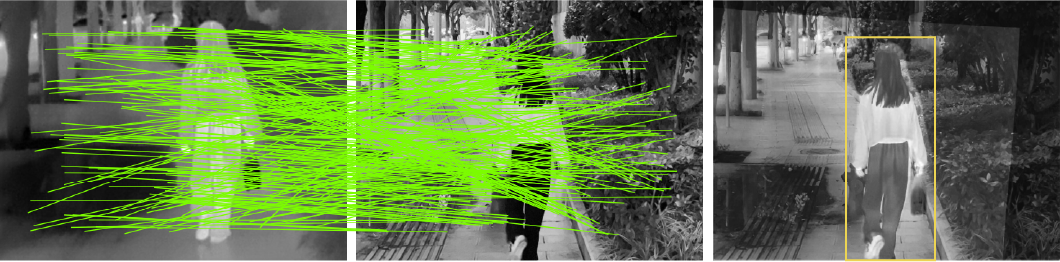}
                \label{chutian1}
        \end{minipage}
        \begin{minipage}{0.31\linewidth}
        \begin{center}
        \footnotesize{\textsf{R2D2}\cite{revaud2019r2d2}}
        \end{center}    
                \centering
                \includegraphics[width=1\linewidth]{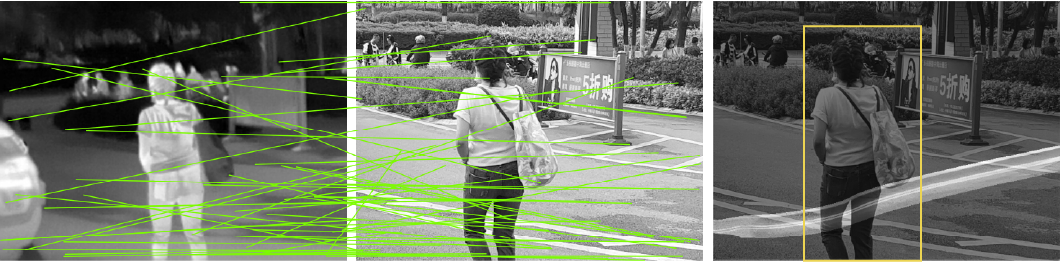}
                \includegraphics[width=1\linewidth]{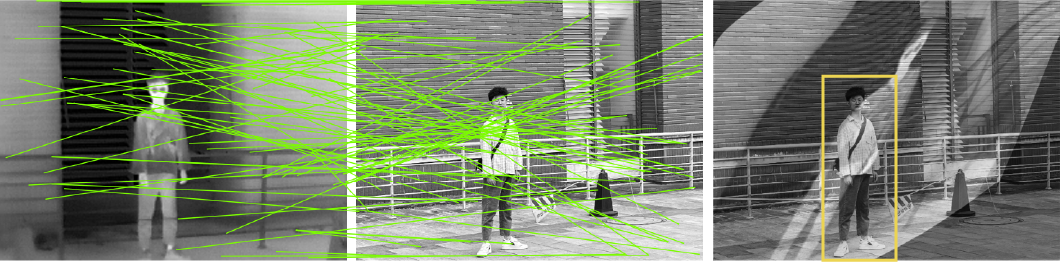}
                \includegraphics[width=1\linewidth]{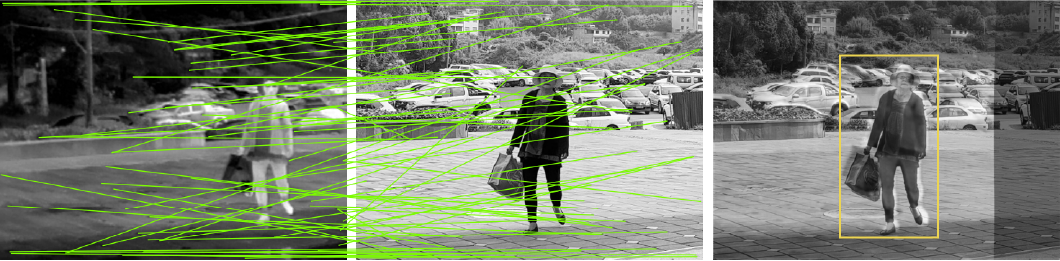}
                \includegraphics[width=1\linewidth]{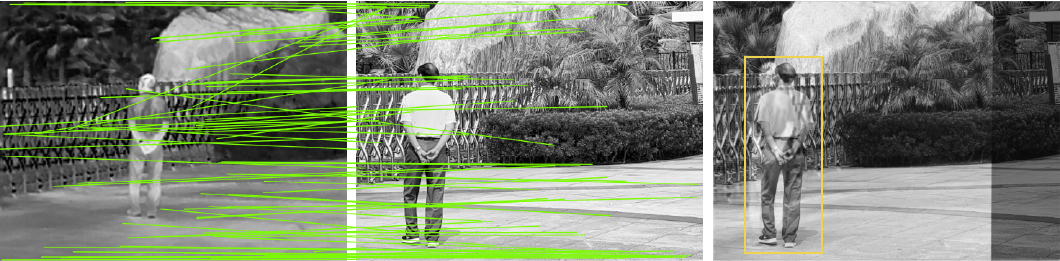}
                \includegraphics[width=1\linewidth]{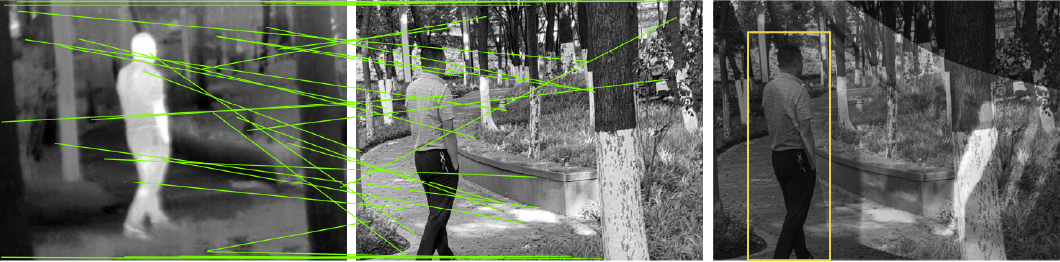}
                \includegraphics[width=1\linewidth]{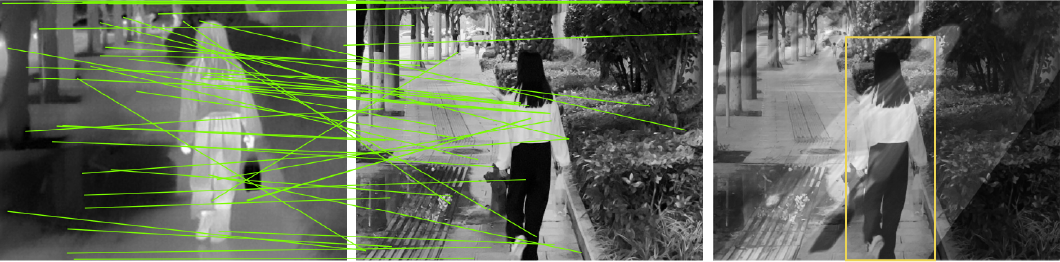}
                \label{chutian1}
        \end{minipage}
        \begin{minipage}{0.31\linewidth}
        \begin{center}
        \footnotesize{\textsf{SP\cite{detone2018superpoint}+SG\cite{sarlin2020superglue}}}
        \end{center}    
                \centering
                \includegraphics[width=1\linewidth]{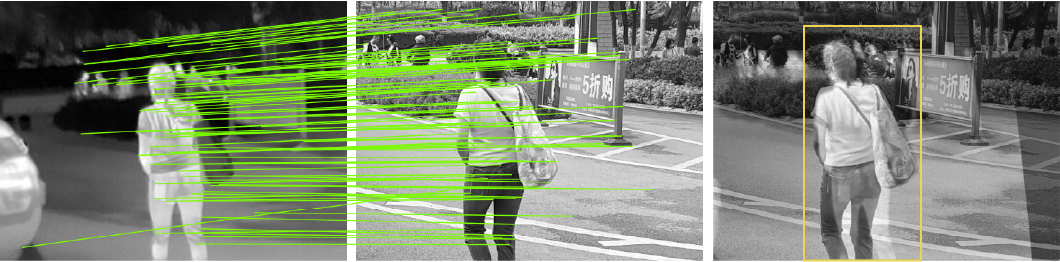}
                \includegraphics[width=1\linewidth]{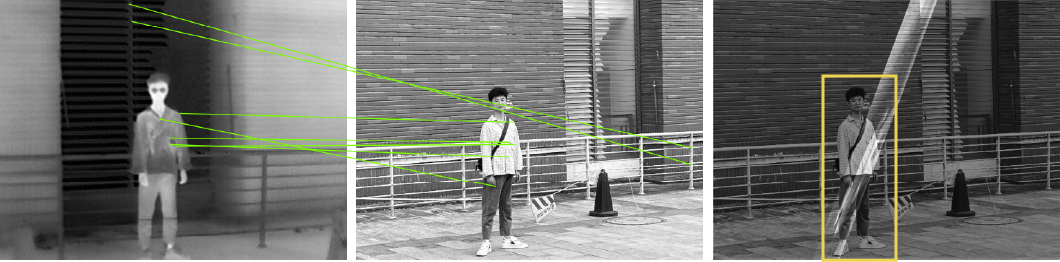}
                \includegraphics[width=1\linewidth]{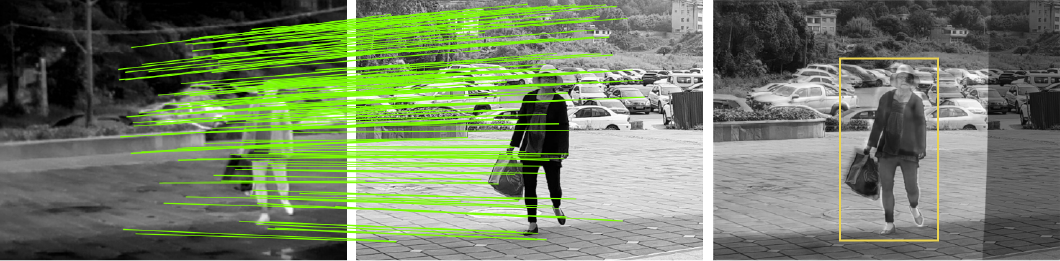}
                \includegraphics[width=1\linewidth]{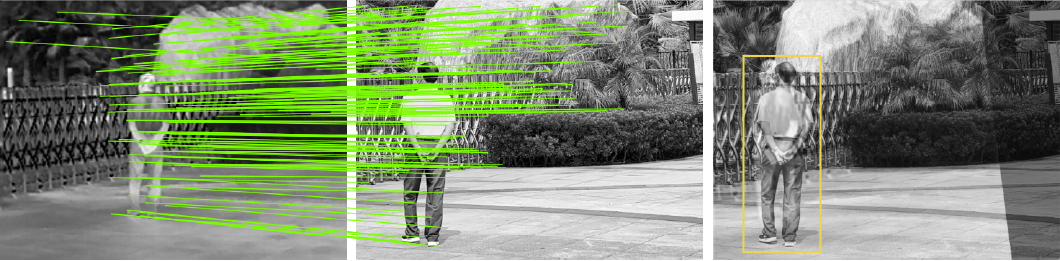}
                \includegraphics[width=1\linewidth]{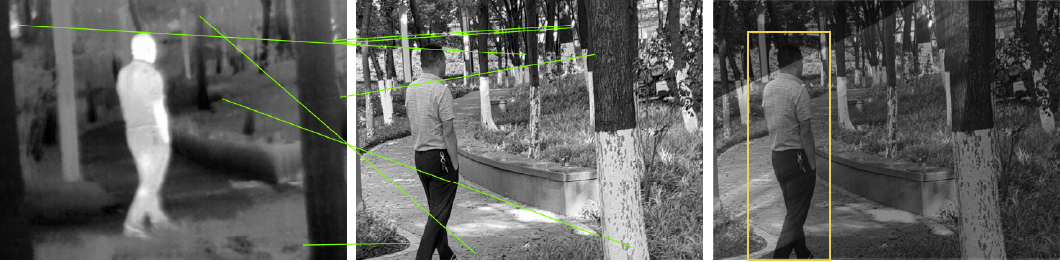}
                \includegraphics[width=1\linewidth]{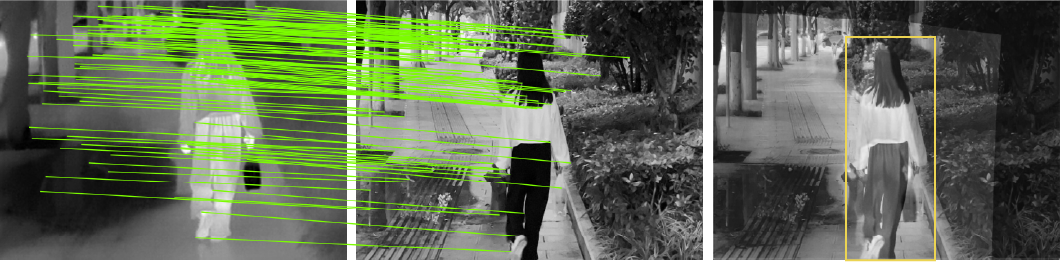}
                \label{chutian1}
        \end{minipage}
        }
        \vspace{-0.3cm}
   
         \subfloat{
        \begin{minipage}{0.31\linewidth}
        \begin{center}
        \footnotesize{\textsf{LoFTR}\cite{sun2021loftr}}
        \end{center}    
                \centering
                \includegraphics[width=1\linewidth]{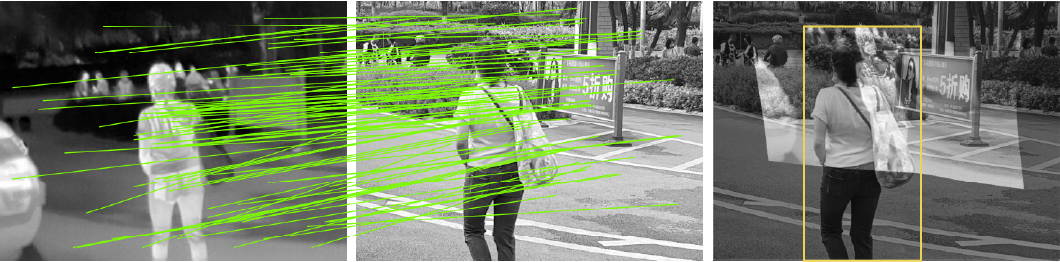}
                \includegraphics[width=1\linewidth]{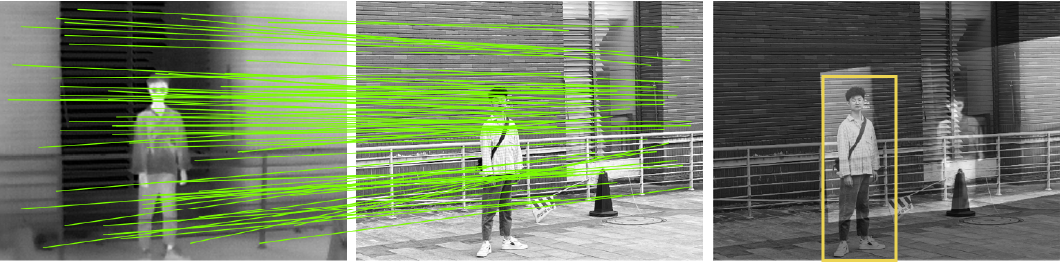}
                \includegraphics[width=1\linewidth]{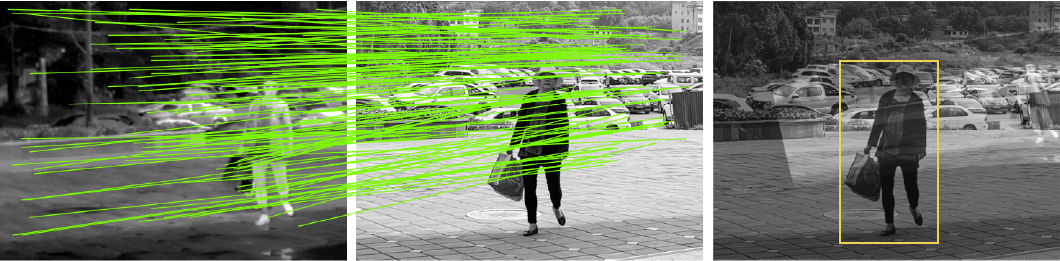}
                \includegraphics[width=1\linewidth]{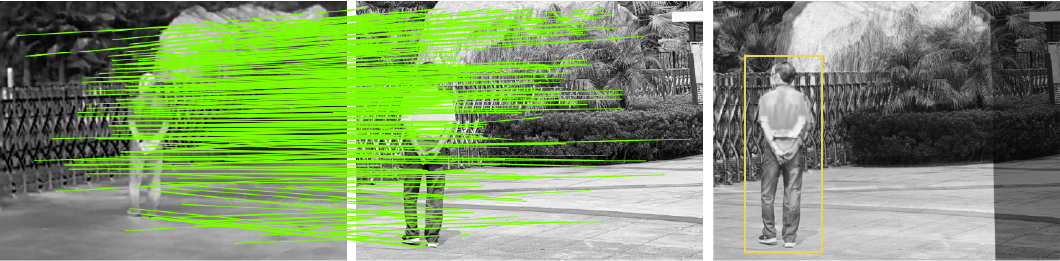}
                \includegraphics[width=1\linewidth]{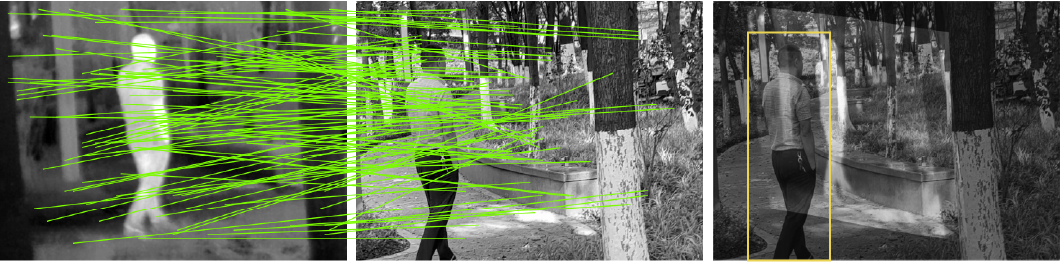}
                \includegraphics[width=1\linewidth]{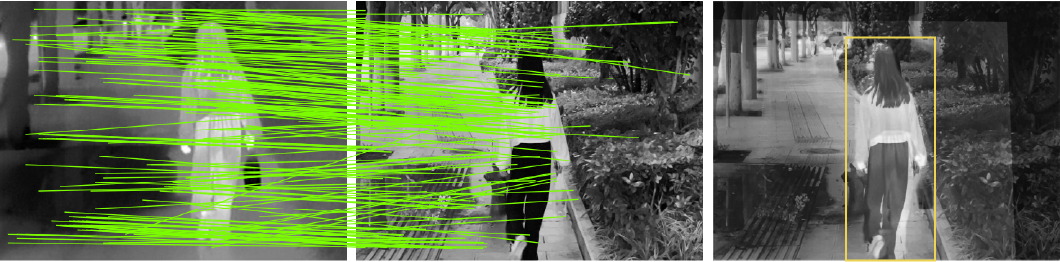}
                \label{chutian1}
        \end{minipage}
       \begin{minipage}{0.31\linewidth}
        \begin{center}
        \footnotesize{\textsf{MatchFormer}\cite{wang2022matchformer}}
        \end{center}    
                \centering
               \includegraphics[width=1\linewidth]{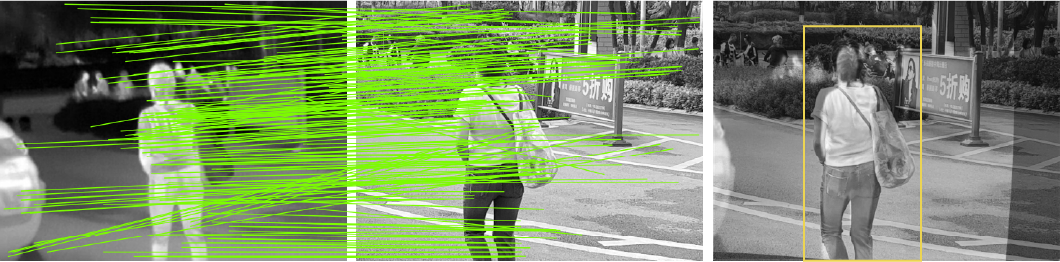}
                \includegraphics[width=1\linewidth]{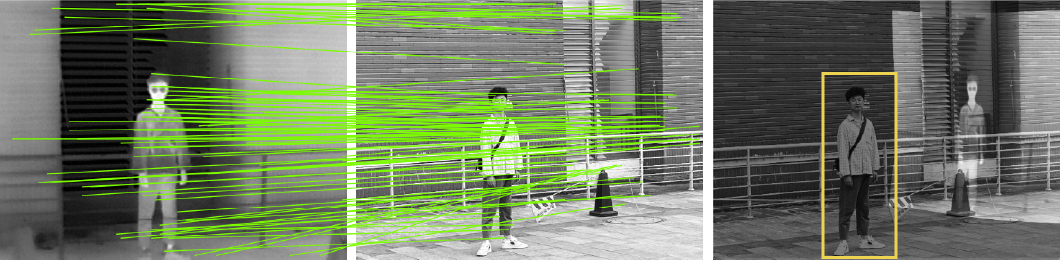}
                \includegraphics[width=1\linewidth]{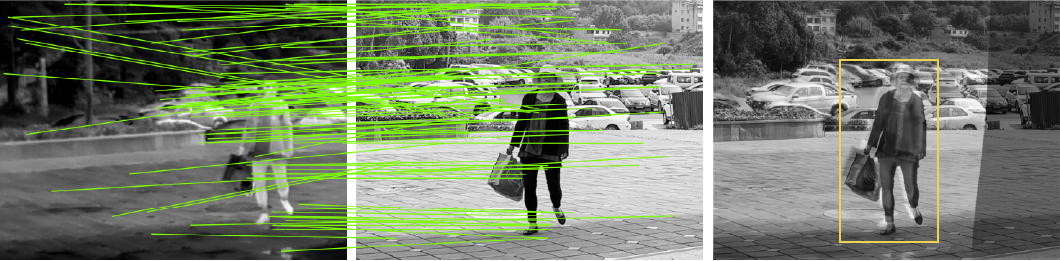}
                \includegraphics[width=1\linewidth]{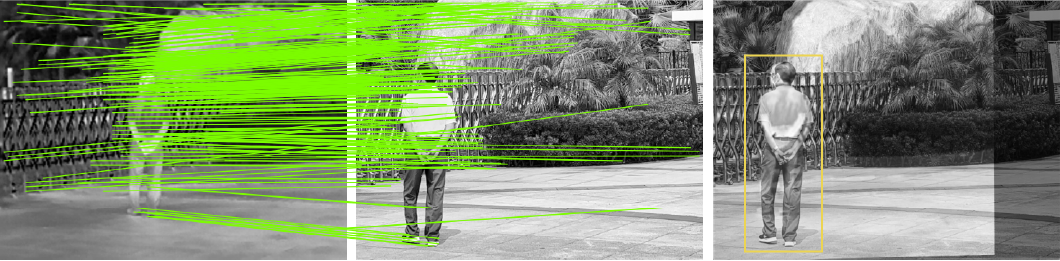}
                \includegraphics[width=1\linewidth]{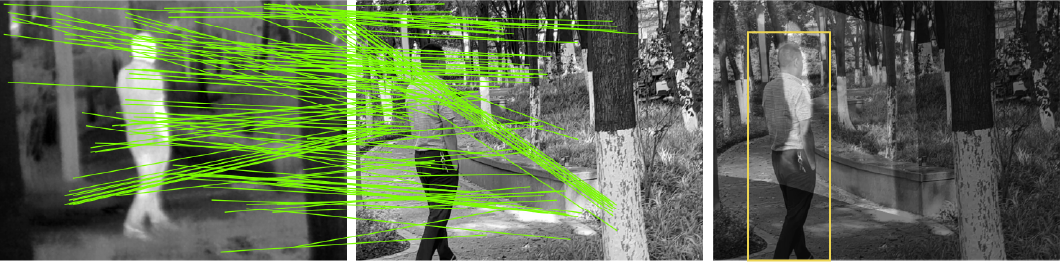}
                \includegraphics[width=1\linewidth]{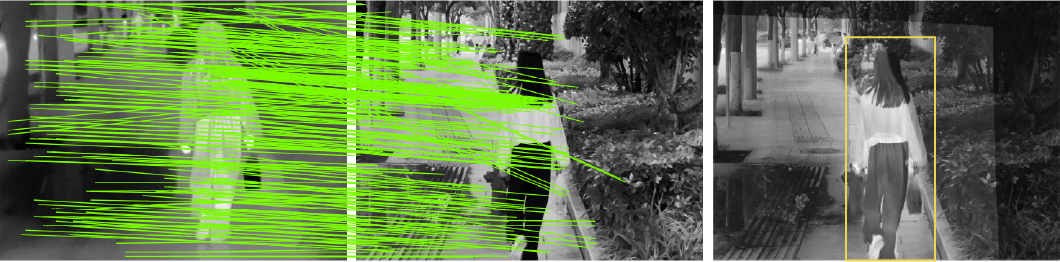}
                \label{chutian1}
        \end{minipage}
        \begin{minipage}{0.31\linewidth}
        \begin{center}
        \footnotesize{\textsf{SA-DNet}}
        \end{center}    
                \centering
                \includegraphics[width=1\linewidth]{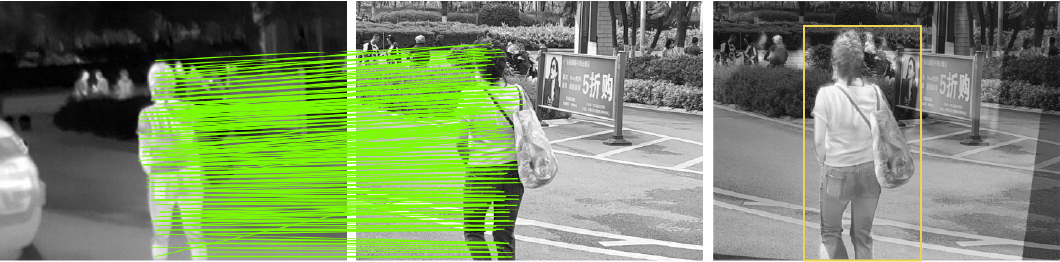}
                \includegraphics[width=1\linewidth]{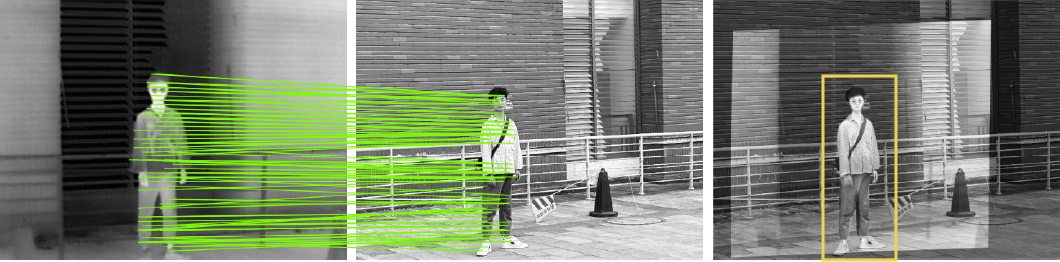}
                \includegraphics[width=1\linewidth]{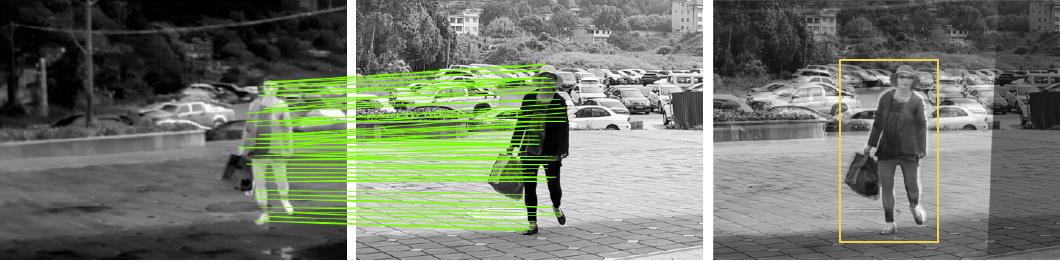}
                \includegraphics[width=1\linewidth]{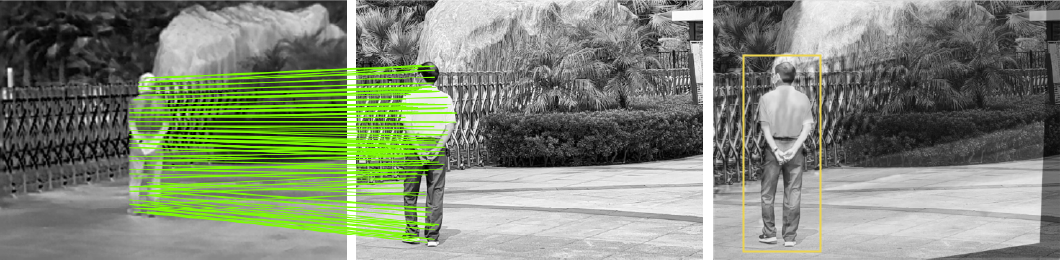}
                \includegraphics[width=1\linewidth]{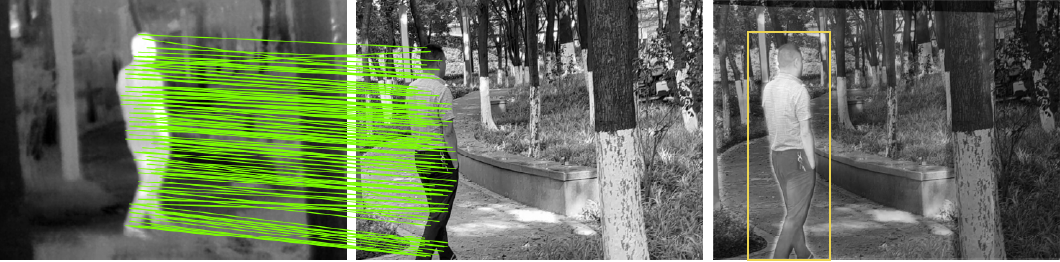}
                \includegraphics[width=1\linewidth]{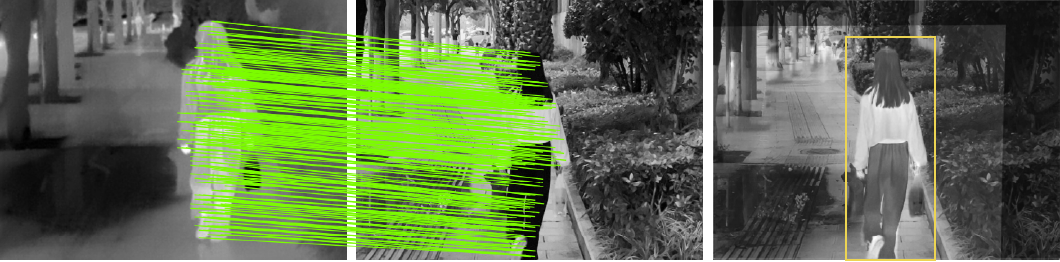}
                \label{chutian1}
        \end{minipage}
        }
        \caption{\textbf{Qualitative results.} Feature matching and fusion results of our SA-DNet and other five feature matching methods in scenes with multiple non-rigid distortions. SA-DNet restricts the feature matching process to pedestrian objects, resists the effect of non-rigid distortions, and achieves better visual results of fused images in pedestrian objects.}
\end{figure*}

\subsection{Applications}
As the main purpose of this paper, semantic-aware on-demand registration aims to provide assistance for advanced vision tasks. Image registration is a precondition for many vision tasks, as a new tpye of registration method, semantic-aware on-demand registration can alter the pre-processing of these tasks. In order to further illustrate the effectiveness of SA-DNet, we evaluate the changes which can be brought by using semantic-aware on-demand registration in image fusion, object detection, and Structure from Motion (SfM) tasks.
\begin{figure*}[htbp]
\centering
\includegraphics[width=180mm]{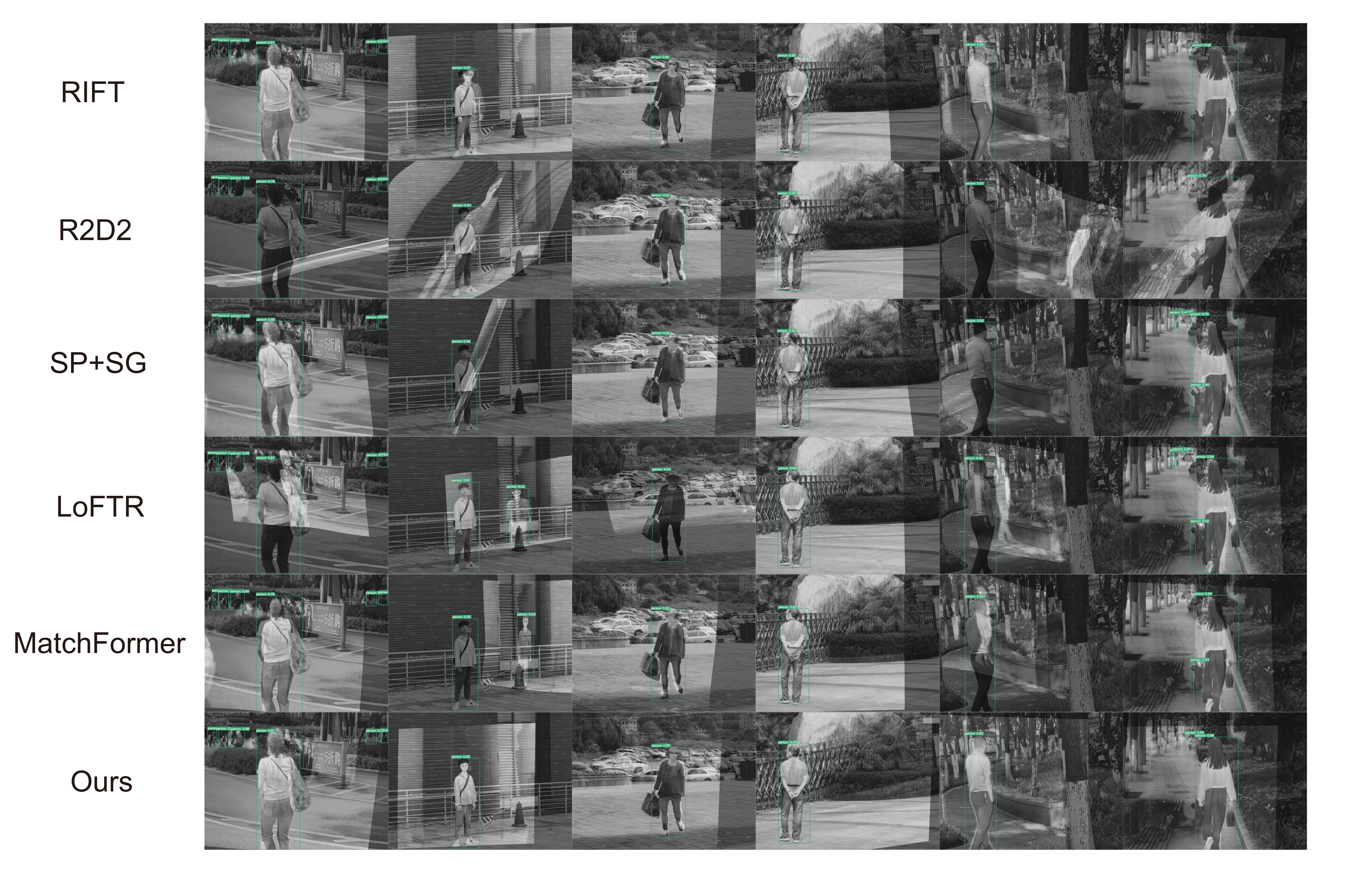}
\caption{\textbf{Object detection of fused images.} Different feature matching methods generate different quality fusion images, which affects the result of object detection.}
\label{fig2:env}
\end{figure*}
\begin{figure*}[htbp]
\centering
\includegraphics[width=180mm]{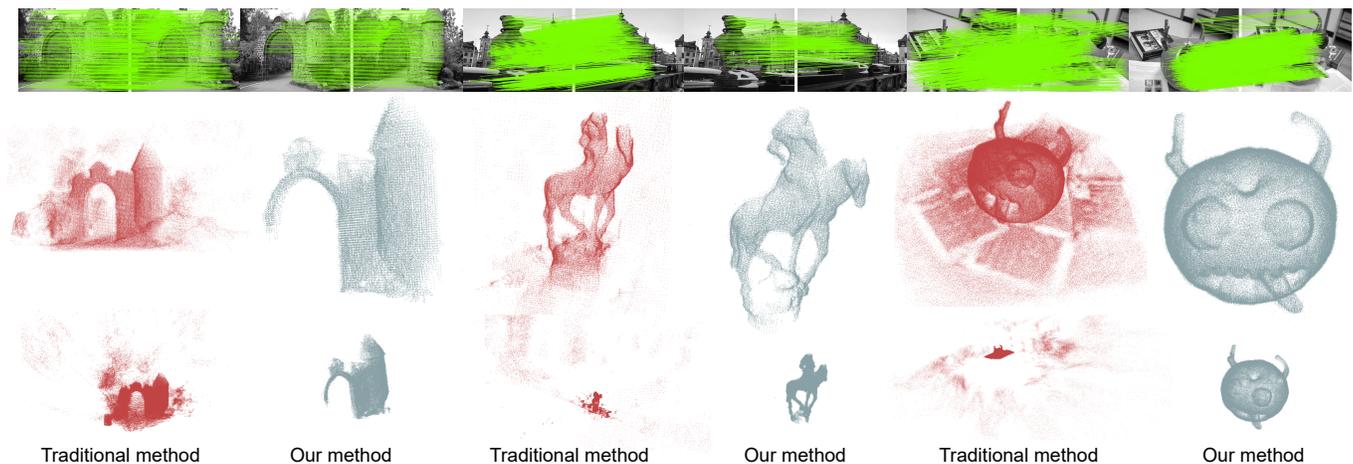}
\caption{\textbf{Experimental results of Structure from Motion (SfM).} The top images represent the reconstructed scenes, we select specific objects in different scenes for feature matching to generate sparse point clouds with different representations. The traditionnal method generates a large number of scene point clouds unrelated to the objects, while our method only generates point clouds containing the objects.}
\label{fig2:env}
\end{figure*}

\textbf{Results on Image Fusion. }We qualitatively and quantitatively evaluate the performance of the fused images after registration using different feature matching methods. Since there is no publicly available complex scene dataset to evaluate for this part of the experiment, we collected 10 pairs of infrared and visible non-registered data in different scenarios, which can be found in the test set of IVS dataset. In order to make a fairly comparison, we replace the corresponding feature points obtained by different feature matching methods with those obtained by our method, and the rest of the network is kept the same, so that different feature matching methods can perform image fusion in the same way.

The pedestrian fraction of the fused image is subjected to an image quality assessment (IQA). Since the quality of the fused image is dependent on the precision of registration, IQA measures registration precision indirectly. The evaluation metrics include average gradient (AG)\cite{cui2015detail}, cross entropy (CE), edge strength (EI)\cite{rajalingam2018hybrid}, mutual information (MI)\cite{qu2002information}, structural similarity (SSIM)\cite{wang2004image}, contrast (CT)\cite{tamura1978textural}. It is worth noting that the source and reference images of cross entropy, mutual information, and structural similarity are the transformed infrared and visible images, respectively. The reason we take this approach is that the increase of semantic object registration error will lead to the difference of corresponding spatial position information in the images, and then the performance of the registration can be reflected by these evaluation method. 

The experimental results are shown in Fig. 10 and Table. III. R2D2 tends to extract highly reliable and repeatable feature points in the images, but the severe nonlinear intensity differences between infrared and visible images cause a great disturbance to such feature point selection strategy, which makes R2D2 get many mismatched points. RIFT maintains good robustness in this part of experiment, but its fused images still have some artifacts from registration errors. SP+SG, LoFTR and MatchFormer obtain many reliable corresponding feature points in different scenes. However, in scenes with a significant non-rigid distortion, a single transformation model is not sufficient to describe all their correspondences. In contrast, due to the feature matching only within specific semantic object, SA-DNet achieves the best results in almost all evaluation metrics and has satisfactory fusion visual effect in sROI.

\textbf{Results on Object Detection. }The fused images will be applied to other more advanced vision tasks, such as object detection, object tracking, medical diagnosis, etc, and the accuracy of image registration will directly affect these vision tasks. We use YOLOv5 to perform object detection on the fused images obtained from the above image fusion experiments, with the results shown in Fig. 11. It can be observed that the pedestrians without good registration will have different degrees of artifacts in the fused images, which results in a worse anchor accuracy and reduced detection confidence during detection. In extreme cases when artifacts deviated from the pedestrian, it even occurs that two identical pedestrians are detected in the same image. SA-DNet, on the other hand, achieves favorable registration on pedestrian objects, allowing the fused images to correctly represent the complementary feature information of infrared and visible images, thereby facilitating the object detection.

\textbf{Results on Structure from Motion (SfM). }Based on a sufficient quantity of 2D images, the SfM calculates the 3D information of 2D images related to their actual scenario in order to assist with various tasks. The classic technique employs SIFT\cite{lowe2004distinctive} to recover the geometric relationships between images as a crucial step in SfM. However, the generated 3D point clouds or models will always include scene information that is irrelevant to the modeling object. If we want to analyze different object point clouds, we need a network specially designed for point clouds. To alleviate this problem, we use the dataset published in\cite{enqvist2011non} for sparse point cloud reconstruction and replace the feature matching step in SfM with our semantic-aware on-demand registration method. As illustrated in Fig. 12, our method provides only sparse point clouds associated with the object, which can be utilized for Object Reconstruction, Virtual Reality (VR), Object Volume Measurement.


\section{Conclusion}
In this paper, we introduce SA-DNet, a network for on-demand semantic object registration. Firstly, we use SAM to perceive semantic regions of interest (sROI) by integrating local and global features with mixed attention, and the HDM performs feature matching on the objects inside the sROI to obtain accurately registered sROI. The experimental results show that SA-DNet has better robustness against non-rigid distortions and we can use different objects as sROI for different needs to achieve customized registration effects. We also demonstrate that semantic-aware on-demand registration can be applied to different vision tasks, altering the processing of these tasks.

In future work, how to further optimize the feature matching effect inside the sROI is the key to improve the robustness of SA-DNet, since as the area of the sROI in the image decreases, the less feature information it contains, which influences the matching effect of SA-DNet in small object. In addition, we intend to develop a faster matcher to improve interoperability with various visual tasks and deploy to a variety of devices.

\bibliographystyle{ieeetr}
\bibliography{myrefs}
\end{document}